\pdfoutput=1

\documentclass[11pt]{article}

\usepackage[final]{acl}

\usepackage{times}
\usepackage{latexsym}

\usepackage[T1]{fontenc}

\usepackage[utf8]{inputenc}

\usepackage{microtype}

\usepackage{inconsolata}

\usepackage{multirow}
\usepackage{booktabs}
\usepackage{array}
\usepackage{enumitem}
\usepackage{xcolor}
\usepackage{colortbl}
\usepackage{amsmath}
\usepackage{amsfonts}
\usepackage{hyperref}
\usepackage{graphicx}
\usepackage{makecell}
\usepackage{ragged2e}
\usepackage{array}
\usepackage{tabularx}

\usepackage{soul}
\definecolor{lightblue}{rgb}{0.9,0.9,1}
\sethlcolor{lightblue}

\newcommand{\tabincell}[2]{\begin{tabular}{@{}#1@{}}#2\end{tabular}}

\title{\textit{Grasping the Essentials}: Tailoring Large Language Models for Zero-Shot Relation Extraction}

\author{
Sizhe Zhou\textsuperscript{1},\quad Yu Meng\textsuperscript{2},\quad Bowen Jin\textsuperscript{1},\quad Jiawei Han\textsuperscript{1} \\
\textsuperscript{1} University of Illinois Urbana-Champaign \quad \textsuperscript{2} University of Virginia  \\
\texttt{\{sizhez, bowenj4, hanj\}@illinois.edu, yumeng5@virginia.edu}
}

\begin{document}
\maketitle

\begin{abstract}
Relation extraction (RE) aims to identify semantic relationships between entities within text. 
Despite considerable advancements, existing models predominantly require extensive annotated training data, which is both costly and labor-intensive to collect. 
Moreover, these models often struggle to adapt to new or unseen relations. 
Few-shot learning, aiming to lessen annotation demands, typically provides incomplete and biased supervision for target relations, leading to degraded and unstable performance.
To accurately and explicitly describe relation semantics while minimizing annotation demands, we explore the \emph{definition only zero-shot RE} setting where \emph{only} relation definitions expressed in natural language are used to train a RE model.
We introduce \textsc{REPaL}, comprising three stages:
(1) We leverage large language models (LLMs) to generate initial seed instances from relation definitions and an unlabeled corpus. 
(2) We fine-tune a bidirectional Small Language Model (SLM) with initial seeds to learn relations for the target domain. 
(3) We expand pattern coverage and mitigate bias from initial seeds by integrating feedback from the SLM’s predictions on the unlabeled corpus and the synthesis history.
To accomplish this, we leverage the multi-turn conversation ability of LLMs to generate new instances in follow-up dialogues, informed by both the feedback and synthesis history.
Studies reveal that definition-oriented seed synthesis enhances pattern coverage 
whereas indiscriminately increasing seed quantity leads to performance saturation.
Experiments on two datasets show \textsc{REPaL} significantly improved cost-effective zero-shot performance by large margins. 
\end{abstract}

\begin{figure}[!t]
    \centering
    \includegraphics[width=\linewidth]{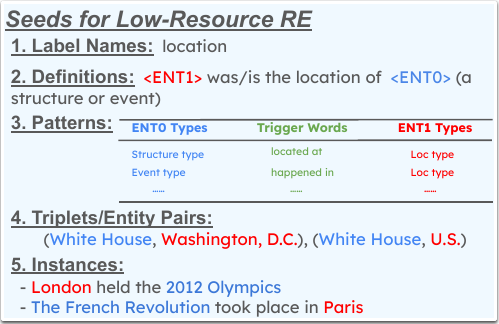}
    \caption{Different types of initial seeds for low-resource RE approaches for example relation P276. It shows using only two instances as seeds fail to cover \textit{structure} type head entities.}
    \label{fig:intro}
    \vspace{-18pt} 
\end{figure}

\section{Introduction}

Relation extraction is a pivotal task in Information Extraction (IE) that involves identifying and classifying semantic relationships between entities within texts. 
It has wide applications in various downstream tasks including knowledge graph construction~\citep{lin2015learning, yu2020relationship}, question answering (QA)~\citep{wang2012relation, wang2016solving} and event mining~\citep{jiao2022open, li-etal-2023-open}.  
Traditional RE works train models with human-labeled data~\citep{han2018hierarchical, han2020more, yamada2020luke}. 
However,  acquiring large-scale, high-quality datasets is challenging and costly in reality. 
To address this data scarcity, few- and zero-shot RE works try to leverage knowledge from LM pre-training or auxiliary sources~\citep{petroni2019language, chia2022relationprompt, han2022ptr, chen2022knowprompt, zhao2023re, zhou2023corpus, wan2023gpt, li-etal-2023-semi, sun2024consistency}.

Despite these advancements, two issues persist in low-resource RE.
The first issue is \emph{the underutilization of relation definitions}.
Relation semantics are generally directional and multifaceted which involve entity-entity interactions and entity-related requirements (see Fig.~\ref{fig:intro}).
Thus, target relation semantics typically can only be partially reflected by most low-resource supervision, such as seed instances, triples, or label names.
Such relation semantic complexity requires detailed elaborations described by relation definitions.
Another issue is the \emph{underutilization of LLMs for zero-shot RE}.
Most LLMs are designed to perform multi-turn conversations and excel in seeking feedback from the dialogue history. 
Such a feature has shown great potential in knowledge-intensive or complex reasoning question-answering tasks~\citep{trivedi2022interleaving, zhou-etal-2023-towards, zhou2023thread}.
Nevertheless, LLM-based low-resource RE works typically rely on single-turn usages.

To address the first issue, this work introduces a new zero-shot RE setting where only relation definitions, instead of seen instances, are provided.
In addition to the fact that relation definitions serve as more precise and less biased initial seeds, such a task setup is realistic as: 
(1) downstream applications such as QA tasks already have explicit definitions of interested relations and obtaining such supervision is generally straightforward; and 
(2) such a setting highlights the importance for RE systems to continuously adapt to new relation types based on corresponding definitions without maintaining a large amount of seen instances and re-training models.

To address the second issue, we propose a novel zero-shot RE framework, \textsc{REPaL}.
\textsc{REPaL} initiates by prompting LLMs to generate positive instances based on predefined definitions and samples negative instances from an unlabeled corpus, forming an initial training set.
This set is then used to train an SLM for inference efficiency and performance. 
Secondly, \textsc{REPaL} acquires and incorporates feedback to address coverage and bias issues from instance generation and SLM training.
For robustness, the feedback consists of two independent components: 
the synthesis dialogues and sampled SLM's inference results on a large unlabeled corpus.
The feedback is utilized to: (1) leverage LLMs' multi-turn conversational ability to recognize the pattern coverage bias, synthesis error, and then generate instances with new or rectified positive patterns, 
and (2) leverage LLMs' reasoning ability to diagnose the SLM's bias and further generate targeted or near-miss negative instances to rectify such bias by explicitly deriving negative definitions.
The whole framework performs iterative refinements in which more and better-quality relation instances are accumulated to improve the task-specialized RE model.

Our data and codes are available here\footnote{https://github.com/KevinSRR/REPaL} and our contributions are as follows: 
\begin{itemize}[leftmargin=*,nosep]
    \item We demonstrate the partial coverage issue of few-shot RE's initial seeds. 
    Our studied definition-oriented RE setting can seamlessly leverage few-shot supervision
    for better pattern coverage and better performance by definition derivation and instance augmentation.
    \item We propose a novel zero-shot RE framework, \textsc{REPaL}, that only requires relation definitions and an unlabeled corpus.
    \textsc{REPaL} iteratively synthesizes both positive and negative instances to enhance pattern coverage and addresses biases by automatically mining and reflecting on feedback from multiple sources, leveraging  the multi-turn conversation capability of LLMs.
    \item Extensive quantitative and qualitative experiments demonstrate the effectiveness and the potential of our task setup and framework.
\end{itemize}

\section{Background}
\begin{figure}[!ht]
    \centering
    \includegraphics[width=0.8\linewidth]{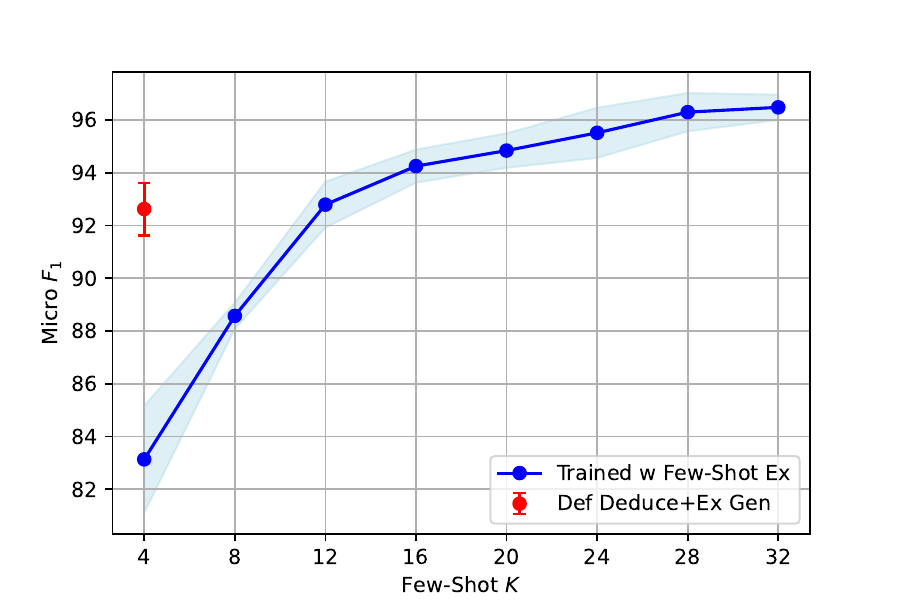}
    \caption{Micro $F_1$ (\%) score of model trained with few-shot instances (\textit{Trained w Few-Shot Ex}) and model trained with instances from our relation definition derivation and instance generation approach (\textit{Def Deduce+Ex Gen}). The error bar/band denotes averaged value $\pm$ standard deviation.}
    \label{fig:def_derive_fewrel_micro_f1}
    \vspace{-12pt}
\end{figure}

\subsection{Definition Only Zero-Shot Relation Extraction Task}

\noindent\textbf{Task Definition 2.1. Definition Only Zero-Shot Relation Extraction.}
We assume that, for any target relation $r(E_0, E_1) \in \mathcal{R}(E_0, E_1)$\footnote{This work addresses sentence-level binary relation extraction, where each instance involves evaluating the relationship between two specific entity mentions.}, only one associated relation definition $d(E_0, E_1)$ is given. 
Here $\mathcal{R}(E_0, E_1)$ denotes the whole binary relation space and $d(E_0, E_1)$ can be a single sentence or a document specifying the target relation $r(E_0, E_1)$. 
$E_0$ and $E_1$ are two entity placeholders.

\begin{figure*}[!ht]
    \centering
    \includegraphics[width=0.8\linewidth]{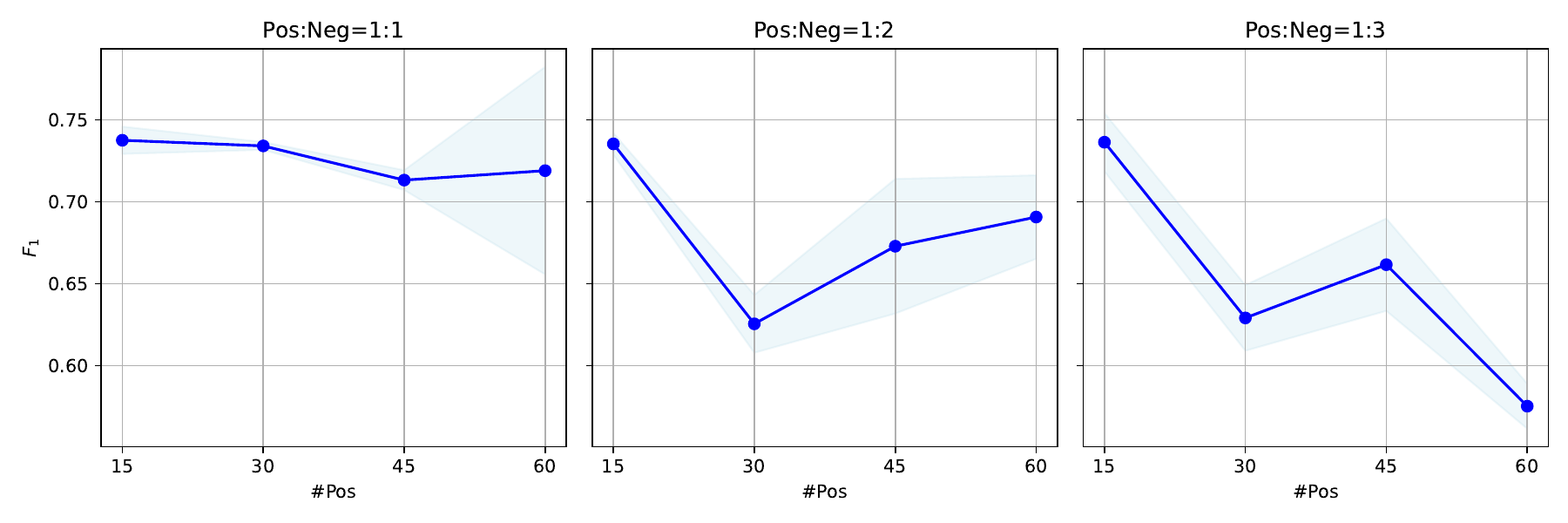}
    \caption{$F_1$ (\%) scores for different setups on the number and ratio of training instances.}
    \label{fig:num_ratio_init_seeds_f1}
    \vspace{-12pt}
\end{figure*}

The goal of \textit{Definition Only Zero-Shot Relation Extraction} task is to extract all relation instances that belong to target relation $r(E_0, E_1)$ from any given set of relation instances $\{(s^i, e_{0}^i, e_{1}^i)\}_{i=1}^{n}$ in a binary classification manner.
Here $s^i$ denotes the $i$-th context text, while $e_0^i$ and $e_1^i$ denote two entity mentions inside $s^i$.
Compared with the traditional zero-shot RE task settings, our task assumes no seen data but only target relation definitions.
Our task also assumes the unknown negative relation space while mainstream zero-shot RE assumes known information of all test relations. 
This makes our task more challenging, but aligns better with real-world scenarios.
Further details and discussions are in Appendix~\ref{sec::appendix:Detailed Discussions on Definition Only Zero-Shot Relation Extraction Setup}.

\subsection{From Few-Shot to Definition-Oriented Low-Resource RE}\label{sec::Background:Derive Relation Definition from Few-Shot Instances}

In this section, we reveal the partial relation pattern coverage issue of commonly adopted RE few-shot setup and further show that the few-shot setup can be converted to our definition-oriented setup while achieving much better results.
We take \textsc{KnowPrompt} \citep{DBLP:conf/www/ChenZXDYTHSC22} as the underlying N-way classification model where N equals the number of test relations. 
It's a prompt tuning model with robust and strong few-shot performance and does not require hand-crafted prompts. 

We experiment on two groups: (1) \textit{Trained w Few-Shot Ex} (baseline group) which has \textsc{KnowPrompt} trained on few-shot instances for evaluation, and (2) \textit{Def Deduce+Ex Gen} which uses GPT-4 to derive each relation's definition given 4-shot instances, generate 15 new instances based on the definition, and train \textsc{KnowPrompt} for evaluation. 

The LLM derived relation definitions compared with original relation definitions are shown in Appendix~\ref{sec::appendix:Relation Definitions Derived by LLM From Few-Shot Instances} and the experiment results are shown in Fig.~\ref{fig:def_derive_fewrel_micro_f1} and Fig.~\ref{fig:def_derive_fewrel_macro_f1} (in Appendix~\ref{sec::appendix:Macro F1 Scores of Few-Shot Method against Definition-Based Method}).
The derived relation definitions show that LLM is capable of deducing suitable yet generalizable relation definitions based on few-shot instances.
However, the coverage of derived definitions is limited by the coverage of few-shot instances. 
This is in accordance with our motivation for definition-based low-resource RE setup. 

Fig.~\ref{fig:def_derive_fewrel_micro_f1} and Fig.~\ref{fig:def_derive_fewrel_macro_f1} show that our definition derivation and instance generation approach achieves much better performance than the model trained only on few-shot instances.
This indicates the approach extends the relation patterns conveyed by the few-shot instances. 
However, we can see our (15 generated + 4 gold shots) trained model has slightly lower performance than 16 gold shots trained model which is due to the partial coverage of relation semantics conveyed by the 4 gold shots instances.
This further illustrates the importance of capturing actual relation definitions instead of few-shot data for low-resource RE approaches. 

\begin{figure*}[!ht]
    \centering
    \includegraphics[width=\linewidth]{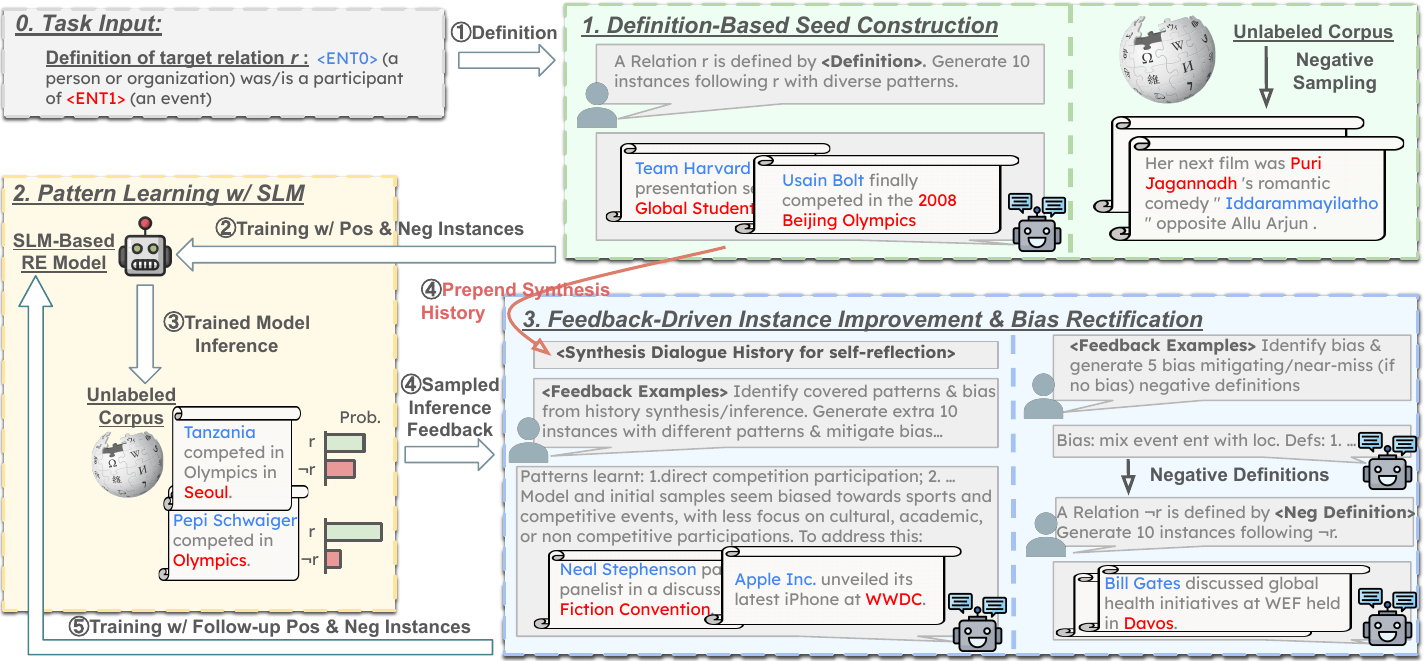}
    \caption{\textsc{REPaL} framework. The trained SLM-Based RE Model is used in inference stage.}
    \label{fig:framework}
    \vspace{-12pt}
\end{figure*}

\subsection{Effect of More Initial Seeds}\label{sec::Background:effect of more initial seeds}
As recent LLMs have enabled larger context windows, a na\"{i}ve method for improving framework performance is to directly query LLMs for more generated instances.  
So we conduct such trials on our definition only zero-shot RE task (Sec. \ref{sec::methodology:def-based initial seed construction}) to investigate whether more initially generated seeds can bring more benefits.
The experiments are based on one split of the DefOn-FewRel and leverage GPT-4 as synthesis backbone with 3 random seeds. 
The positive instances are generated with prompt templates shown in Table~\ref{tab:prompt templates for initial seed construction with LLM_brief}, Table~\ref{tab:prompt templates for initial seed construction with LLM_medium}, and Table~\ref{tab:prompt templates for initial seed construction with LLM_implicit} while the negative instances are randomly sampled from the unlabeled corpus. 
The instances are used to train a RE model adapted from \texttt{roberta-large-mnli}~\cite{liu2019roberta}. 
The quantitative results are shown in Fig.~\ref{fig:num_ratio_init_seeds_prec}, Fig.~\ref{fig:num_ratio_init_seeds_recall}, and Fig.~\ref{fig:num_ratio_init_seeds_f1}. 
We can see that synthetic data by LLM is generally beneficial, but generating more initial seeds does not guarantee better results. 
Larger p\&n\footnote{We abbreviate the number of positive instances for each target relation as p and the number of negative instances as n.} or larger n:p ratio both lead to higher precision and lower recall.

One explanation for these trends is that more positive seeds does not bring larger positive pattern coverage but results in over-fitting. 
However, more randomly sampled negative seeds lead to larger negative pattern coverage.
Therefore the model tends to give positive predictions on instances with some dominant positive patterns but identify other minority patterns as negative. 
Note that when n\&p are small, usually recall is pretty high while precision is low. 
This may also be caused by our task assumption on unknown negative relation space. 
The above observations and analysis motivates our design of a feedback-driven follow-up generation and refinement paradigm for instance synthesis.

\section{Methodology}
Our proposed \textsc{REPaL} is an iterative framework which consists of three major components: 
(1) Definition-based seed construction with LLM and the unlabeled corpus (Sec.~\ref{sec::methodology:def-based initial seed construction}). 
(2) Pattern learning with RE-specialized SLM (Sec.~\ref{sec::methodology:pattern learning with SLM}) which trains a SLM extractor with initial seeds for inference efficiency and performance.
(3) Feedback-driven instance improvement and bias rectification (Sec.~\ref{sec::methodology:feedback-driven instance improve and bias rectify}) by providing LLMs with independent feedback and leveraging LLMs' multi-turn conversations for followup positive and negative instance synthesis. 
The framework overview is shown in Fig.~\ref{fig:framework}.

\subsection{Definition-Based Seed Construction with LLM and Unlabeled Corpus}\label{sec::methodology:def-based initial seed construction}
LLMs, such as the GPT family, are pre-trained for the domain adaptation ability \citep{radford2019language}.
They have demonstrated to contain factual relation knowledge \citep{petroni2019language} and are widely used as data generator for various tasks such as text classification \citep{meng2022generating} and dialogue systems \citep{abdullin2024synthetic}. 
Follow-up evaluation studies have shown that LLMs are relatively skilled at constrained content generation, storytelling, and rationale generation \citep{sun-etal-2023-evaluating, keskar2019ctrl}.
Following such observations as well as the goal to tackle the data scarcity issue of zero-shot RE, \textsc{REPaL} first leverages LLMs to generate $K_{p_0}$ positive seeds, $\{(s_p^i, e_{0, p}^i, e_{1, p}^i)\}_{i=1}^{K_{p_0}}$, conditioned on the target relation definition $r(E_0, E_1)$. 
We carefully design three prompts for this step (shown in Appendix~\ref{sec::appendix:Prompt Templates for Definition-Based Seed Construction with LLM}) to encourage the LLM to generate comprehensive patterns from three perspectives in terms of complexity: \texttt{brief}, \texttt{medium}, and \texttt{implicit}. 
Based on exploratory experiments, they yield different types of relation patterns in accordance with our design expectations.
Details of experiments and quantitatively how choice of different prompts affect the results are in Sec.~\ref{sec::results and analysis:Effect of Positive Templates}.

In addition to positive seed generation, constructing negative seeds is also vital for better extraction for the target relation $r(E_0, E_1)$, as our task setting assumes no prior knowledge on the negative relation space in order to mimic the real-life deployments~\citep{li2022open}. 
Our negative seed construction is based on the hypothesis that, with a normal large-scale unlabeled corpus, the proportion of target relation instances is relatively small. 
Thus \textsc{REPaL} conducts random sampling over the unlabeled corpus, which is expected to yield an acceptable set of negative instances denoted as $\{(s_n^i, e_{0, n}^i, e_{1, n}^i)\}_{i=1}^{K_{n_0}}$. 
To address potential issues in extreme cases, \textsc{REPaL} designs countermeasures in Sec.~\ref{sec::methodology:feedback-driven instance improve and bias rectify}.

\subsection{Pattern Learning with RE-Specialized SLM}\label{sec::methodology:pattern learning with SLM}
Given the relatively high computational costs of fine-tuning and inference with LLMs and the limitation of vanilla in-context learning (ICL) due to LLMs' difficulties in fully processing long-context demonstrations~\citep{Ma2023LargeLM}, we leverage SLMs as tunable and task-specialized extractors. 
SLMs also enable faster inference on massive unlabeled corpus to mine feedback in Sec.~\ref{sec::methodology:feedback-driven instance improve and bias rectify}.
Specifically, we adapt a natural language inference (NLI) model~\citep{obamuyide2018zero, liu2019roberta, sainz2021label} to be relation-specific binary classifiers for simplicity and leave other architectures or LLMs for future research.

For each training instance $(s^j, e_{0}^j, e_{1}^j)$, the input is formulated as in NLI style:
\vspace{-6pt}
\begin{align*}
    \texttt{Premise}_j :&= s^j, \\
    \texttt{Hypothesis}_j :&= d(E_0=e_0^j, E_1=e_1^j).
\end{align*}
Given a SLM model $\mathcal{M}$, we obtain the encoded sequence hidden states $\mathbf{H}$ by: 
\vspace{-6pt}
\begin{equation*}
    \mathbf{H} = \mathcal{M}(\texttt{Premise}_j\, \texttt{[SEP][SEP]}\, \texttt{Hypothesis}_j)
\end{equation*}
and the NLI logits $\mathbf{z}=[z_E, z_N, z_C] \in \mathbb{R}^3$ is computed as:
\vspace{-9pt}
\begin{equation*}
    \mathbf{z} = \mathbf{W} \cdot \mathbf{H}_{[CLS]} + \mathbf{b}.
\end{equation*}
Finally, $P_{j}$, the probability of instance $(s^j, e_{0}^j, e_{1}^j)$ following relation $r(E_0, E_1)$, is computed as the normalized logit of \texttt{ENTAILMENT} label:
\begin{equation*}
    P_{j} =  \frac{e^{{z}_{E}}}{\sum_{c \in \{C, N, E\}} e^{{z}_c}} ,
\end{equation*}
where $C, N, E$ denote logits $\mathbf{z}$'s indices for NLI label \texttt{CONTRADICTION}, \texttt{NEUTRAL}, \texttt{ENTAILMENT} respectively.
The binary classification loss \citep{shannon1948mathematical} for relation $r(E_0, E_1)$ is defined as:
\begin{equation}\label{eq:binary classification train loss}
    \begin{split}
        \mathcal{L} = - \frac{1}{|B|} \sum_{\substack{(s^j, e_{0}^j, e_{1}^j) \in B}} &[ y_j \log (P_{j}) \\
        + &(1-y_j)  \log (1- P_j) ] ,
    \end{split}
\end{equation}
where $B$ denotes the batched data sampled from $\{(s_p^i, e_{0, p}^i, e_{1, p}^i)\}_{i=1}^{K_{p_0}} \bigcup \{(s_n^i, e_{0, n}^i, e_{1, n}^i)\}_{i=1}^{K_{n_0}}$.

\subsection{Feedback-Driven Instance Improvement and Bias Rectification}\label{sec::methodology:feedback-driven instance improve and bias rectify}

After obtaining the trained SLM relation extractor, \textsc{REPaL} further introduces a follow-up feedback-driven iterative refinement approach with intuitions listed as follows: 
(1) The initially generated positive seeds might only have a partial relation pattern coverage or concentrate on common patterns due to LLM's longtail deficiency~\citep{mallen2022not}. 
Instead of one-time generation, iterative follow-up generations by incorporating feedback from examining previously generated results are better for pattern coverage.
(2) Bias or errors can be introduced by previous two steps (Sec.~\ref{sec::methodology:def-based initial seed construction} and \ref{sec::methodology:pattern learning with SLM}). 
One potential source of bias is the generated positive seeds or the randomly sampled negative seeds.  
Another potential source of bias is the randomness of SLM training over limited training samples (see Fig.~\ref{fig:case study for pos gen} and Fig.~\ref{fig:case study for neg def gen}). 
Identifying and rectifying bias requires a feedback-driven follow-up refinement approach. 

Specifically, we first leverage the trained SLM extractor to obtain the inference results on the unlabeled corpus.  
The first component of the feedback is constructed by randomly sampling instances within a predicted probability range (see Appendix \ref{sec::appendix:implementation details} for details). 
It is constructed for follow-up positive instance and follow-up negative instance generation respectively, differing only on the probability ranges.

The other component of the feedback is LLMs' synthesis history. 
For follow-up positive instance generation, we leverage the multi-turn conversational feature of LLMs where the positive instance synthesis history is prepended to a new turn of conversation asking for positive instances with different patterns. 
The SLM's inference feedback is integrated into the new turn of conversation so that LLMs can identify the correct and incorrect patterns learned by SLMs.
The double channel feedback design, incorporating both SLM inference and synthesis history, enhances the robustness of this step as the followup synthesis can still rely on LLMs' self-reflection if the inference feedback is uninformative.

For follow-up negative instance generation, we divide it into two sub-steps following the divide-and-conquer philosophy: the first sub-step is feedback-driven negative relation definition generation and the subsequent sub-step is negative instance generation based on the negative relation definitions. 
LLM is queried to examine the SLM's inference feedback and identify the incorrect patterns learned by SLM so that newly generated negative relations can address such patterns. 
Similarly, for robustness in cases where SLM's inference feedback is uninformative, we query LLM to generate near-miss negative relation definitions based on the positive relation definition to better distinguish hard cases. 
After acquiring the negative relation definitions, the second sub-step is the same as the initial positive seed generation process in Sec.~\ref{sec::methodology:def-based initial seed construction}. 
The prompts used in this section are placed in Appendix~\ref{sec::appendix:Prompt Templates Used in Feedback-Driven Instance improvement and Bias Rectification}.
After obtaining all the follow-up positive and negative instances, we iteratively train the SLM extractor.

\section{Experiments}
\subsection{Evaluation Dataset Construction}\label{sec::experiments:evaluation dataset construction}

\begin{table*}[!t]
    \small 
    \centering
    \begin{tabular}{lcccccccc}
        \toprule 
        \multirow{2}{*}{Model} & \multicolumn{4}{c}{DefOn-FewRel} & \multicolumn{4}{c}{DefOn-Wiki-ZSL}\\
        \cmidrule(l){2-5}
        \cmidrule(l){6-9}
        & Precision & Recall & F$_1$ & Macro-F$_1$ & Precision & Recall & F$_1$ & Macro-F$_1$ \\
        \midrule
        \textit{Fully-Supervised} \\
        \quad   \textsc{RoBERTa NLI}  & 79.36 & 98.46 & 86.99 & - & 68.66 & 97.14 & 78.46 & -\\
        \midrule
        \textit{Zero-Shot} \\ 
        \quad   \textsc{Random Guess} & 7.14 & 50.77 & 12.52 & - & 6.67 & 51.01 & 11.67 & - \\
        \quad   \textsc{GPT-3.5} & 55.09 & 61.81 & 53.94 & - & 42.64 & 45.70 & 39.60 & - \\
        \quad   \textsc{RE as QA}  & 84.57 & 73.42 & 74.34  & 84.67 & 75.72 & 61.48 & \textbf{60.89} & 70.27 \\
        \quad   \textsc{RoBERTa NLI}  & 48.79 & 81.17 & 55.07 & - & 36.23 & 62.65 & 35.28 & - \\
        \quad   \textsc{ZS-BERT}  & 35.91 & 35.70 & - & 35.78 & 37.11 & 33.85 & - & 35.26 \\
        \quad   \textsc{RelationPrompt}  & 74.39 & 66.89 & - & 67.78 & 71.89 & 60.50 & - & 61.08 \\
        \quad   \textsc{RE-Matching}  & 77.49 & 72.95 & - & 75.11 & 73.10 & 68.99 & - & 70.97 \\
        \rowcolor{gray!20}  \quad   \textbf{\textsc{REPaL} (w GPT-4o mini)}  & 73.27 & 73.75 & 70.46 & 90.01 & 65.67 & 43.53 & 44.78 & 78.47 \\
        \rowcolor{gray!20}  \quad   \textbf{\textsc{REPaL} (w GPT-4o)}  & 78.86 & 77.28 & \textbf{74.61} & \textbf{91.71} & 68.98 & 47.63 & 47.80 & \textbf{80.96} \\
        \bottomrule
    \end{tabular}
    \caption{Evaluation results of relation extraction models under zero-shot and fully-supervised settings. \textsc{REPaL} is trained on 15p15n initial instances to acquire SLM inference feedback and then trained with additional 15p15n follow-up instances (30p30n in total). We show the results with different backbone synthesis LLMs (GPT-4o and GPT-4o mini). Note that the results for GPT-3.5 and \textsc{RE as QA} baselines on the \textit{Definition Only Zero-Shot Relation Extraction} setting are based on the down-sampled subsets with 30 samples/relation to reduce inference cost.}
    \label{tab:main_eval_results}
    \vspace{-12pt}
\end{table*}

We first construct datasets for model evaluation on the \textit{Definition Only Zero-Shot Relation Extraction} task.
Our constructed datasets are based on FewRel~\citep{han-etal-2018-fewrel}
and Wiki-ZSL~\citep{chen2021zs} respectively\footnote{An ideal test set for our setting should follow: (1) annotations should follow the officially documented relation definitions and the sentences should be sufficient to deduce the target relations without external knowledge; (2) relations are better semantically disjoint without ambiguity so that we do not need to manually adjust binary test labels for overlapped relations in each test iteration. These principles also apply to existing multi-way classification RE datasets. However, we leave these for future dataset related work.}.
The public portion of FewRel contains 80 relations, each comes with 700 instances annotated by crowd workers.
Wiki-ZSL consists of 113 relations with total 93,383 instances obtained by distant supervision.
As their relations are from WikiData's\footnote{\href{https://www.wikidata.org/wiki/Wikidata:Main_Page}{WikiData main page}} properties, we query the property definitions and slightly modify them to become complete sentences with entity placeholders \texttt{[ENT0]} and \texttt{[ENT1]} (corresponding to $E_0$ and $E_1$ as in definition notation $d(E_0, E_1)$). 
For evaluation, we sample 5 groups of 14 FewRel relations and 3 groups of 15 Wiki-ZSL relations as test sets.

To construct unlabeled corpus, for each group of the test set, we down-sample 10,000 instances from the unlabeled corpora used by \citeauthor{gao2020neural} for few-shot relation learning which contains 744 distantly supervised relations (labels are not used in this work) and in total 899,996 instances \citeyearpar{gao2020neural}. 
The final test sets with the corpus are denoted as DefOn-FewRel and Defon-Wiki-ZSL respectively.

\subsection{Baseline Methods}
We compare our method with the following baselines under \textit{Fully-Supervised} and \textit{Zero-Shot} settings:
(1) \textsc{Random Guess}: gives uniformly random binary predictions.
(2) \textsc{GPT-3.5} \citep{brown2020language, ouyang2022training}: uses \texttt{gpt-3.5-turbo-0125} model and formulates RE as a double choice problem, answering whether two entity mentions in each test instance follow the given target relation definition. 
(3) \textsc{RE as QA}: similar to QA4RE \citep{zhang2023aligning}, we design more QA-like prompt templates based on the relation definitions and \texttt{gpt-3.5-turbo-0125} model to formulate RE as a multiple choice problem (double choice for \textit{Definition Only Zero-Shot Relation Extraction} task and multi-choice for traditional zero-shot RE).
(4) \textsc{RoBERTa NLI} \citep{devlin2018bert, liu2019roberta}: our adopted SLM RE backbone model (Sec.~\ref{sec::methodology:pattern learning with SLM}).
It adopts 100 gold positive and 100 gold negative instances for each relation under the \textit{Fully-Supervised} setting. 
(5) \textsc{ZS-BERT} \citep{chen2021zs}: a Siamese-based zero-shot RE framework trained on seen labeled instances and gives prediction by nearest neighbour search comparing encoded sentence representations and relation definition representations.
(6) \textsc{RelationPrompt} \citep{chia2022relationprompt}: a Seq2Seq-based zero-shot RE framework that trains GPT-2 \citep{radford2019language} to generate relation instances conditioned on relation names and train BART \citep{lewis-etal-2020-bart} to generate the extracted relation triple on seen data. 
For unseen data, it fine-tune BART on synthetic data generated by trained GPT and then predicts. 
(7) \textsc{RE-Matching} \citep{zhao2023re}: a Siamese-based zero-shot RE model that encodes entity types and relation definitions for each relation on seen data and conducts nearest neighbour search for prediction on unseen data. 

Note that \textsc{ZS-BERT}, \textsc{RelationPrompt} and \textsc{RE-Matching} require being trained on labeled seen data and their predictions on unseen data need to be conducted in a multi-way classification manner where information of all relations is leveraged. 
These three baselines are trained on the relation instances not in the test set while leaving 5 relations' instances for dev set.

\subsection{Experiment Setup}

\paragraph{Evaluation Methods.}
Following our \textit{Definition Only Zero-Shot Relation Extraction} setting, evaluation is conducted similar to cross validation where each iteration one test relation is treated as the gold positive target relation and remaining test relations serve as gold negative test relations (our setting assumes all negative relations are unknown in terms of relation definitions and any distribution information).
Each test iteration is a binary classification problem with precision, recall and F$_1$ calculated. 
Table \ref{tab:main_eval_results} shows the main results where Precision, Recall and F$_1$ indicates the corresponding binary classification metric scores averaged across all test iterations. 
If the cell for F$_1$ column is empty, then the Precision and Recall represents the Macro Precision and the Macro Recall corresponding to Macro F$_1$.
Since \textsc{ZS-BERT}, \textsc{RelationPrompt} and \textsc{RE-Matching} require the traditional zero-shot RE setting where multi-way classification is assumed, we further train the \textsc{KnowPrompt} \citep{DBLP:conf/www/ChenZXDYTHSC22} multi-way RE classifier with all the positive instances of test relations synthesized by \textsc{REPaL} in the \textit{Definition Only Zero-Shot Relation Extraction} setting for comparison. 
This leads to the Macro-F$_1$ column. 
See Appendix~\ref{sec::appendix:implementation details} for further implementation details
and see Appendix~\ref{sec::appendix:Detailed Discussions on Definition Only Zero-Shot Relation Extraction Setup} for details of our definition only zero-shot RE setting.

\section{Results and Analysis}

\subsection{Main Results}
The main results are shown in Table~\ref{tab:main_eval_results}. 
\textsc{REPaL} achieves generally better performance compared with zero-shot baselines with large margins for both evaluation datasets. 
This shows the effectiveness of our method in low-resource settings and the robustness across different benchmarks. 

The advance of \textsc{REPaL} is slightly lower in DefOn-Wiki-ZSL compared to DefOn-FewRel which is because Wiki-ZSL is much more noisy due to distant supervised annotations. 
By comparing the absolute values of F$_1$ scores derived for our zero-shot RE setting with the Macro-F$_1$ scores derived for traditional zero-shot RE setting, it can be concluded that our definition only zero-shot setting is much more challenging. 
Once we break the assumption of unknown negative relation space, models can take shortcuts to distinguish different relations without actually comprehending the relations.

Compared with fully-supervised baselines, it shows there is still room for improvement.
This is related to our task's assumption on unknown negative test relation space which is normally overlooked in mainstream zero-shot RE works.
Note that, on our \textit{Definition Only Zero-Shot RE setting}, \textsc{RE as QA} achieves better performance on Defon-Wiki-ZSL but lower on traditional zero-shot RE setting.
This shows that LLMs are capable of judging whether an instance follows a certain relation if given a clear relation definition. 
However, they are less competent for dealing with multiple relations. 
Meanwhile, inference with LLMs are much more costly as the \textsc{RE as QA} (w/ GPT-3.5) would cost around \$260 for one DefOn-FewRel split on our setting if without down-sampling.
For traditional setting, the \textsc{RE as QA} would cost around \$2.3.
\textsc{REPaL}, in contrast, even with GPT-4-0125-preview, costs around \$3.7 for generating 30p30n train and 30p30n dev examples\footnote{Approximated by multiplying the number of input and output tokens of GPT-4o with the price of GPT-4-0125-preview.}. 
The generating instances are further reused for both settings. 

\subsection{Ablation Study}\label{sec::results and analysis:ablation study}

\begin{table}[ht]
    \small
    \centering
    \begin{tabular}{lccc}
        \toprule 
        \multirow{2}{*}{Model} & \multicolumn{3}{c}{DefOn-FewRel} \\
        \cmidrule(l){2-4}
        & Precision & Recall & F$_1$ \\
        \midrule
        \textsc{REPaL} & \textbf{78.86} & 77.28 & \textbf{74.61} \\
        \quad -\,N\_feedback & 76.74 & 77.47 & 73.04 \\
        \quad\quad -\,P\_feedback & 73.59 & 75.90 & 70.03 \\
        \quad\quad\quad -\,P\_init & 48.79 & \textbf{81.17} & 55.07 \\
        \bottomrule
    \end{tabular}
    \caption{Ablation results of \textsc{REPaL}. P\_init, P\_feedback, and N\_feedback denote initial positive generation, feedback-driven follow-up positive generation and negative generation respectively. The ablations are based on \textsc{REPaL} w GPT-4o.}
    \label{tab:ablation_fewrel_30p60n}
    \vspace{-10pt}
\end{table}

To investigate the effectiveness of our framework design, we conduct ablation studies with results shown in Table~\ref{tab:ablation_fewrel_30p60n}.
The initial seed generation brings considerable performance advance which shows LLM's power of domain adaptation is fully leveraged given a clear relation definition.  
Compared to the results shown in Sec.~\ref{sec::results and analysis:Effect of Positive Templates} where we generate 30p30n instances all at once, \textsc{REPaL} achieves better recall and F$_1$ both with and without follow-up negative instance generation. 
This indicates the importance of feedback-driven generation design.

Furthermore, the negative follow-up instance generation further boosts the precision, demonstrating its effectiveness on rectifying SLM's bias in distinguishing positive and negative relations.

\subsection{Effect of Positive Templates}\label{sec::results and analysis:Effect of Positive Templates}
\begin{table}[ht]
    \small
    \centering
    \begin{tabular}{lccc}
        \toprule 
        \multirow{2}{*}{Model} & \multicolumn{3}{c}{DefOn-FewRel} \\
        \cmidrule(l){2-4}
        & Precision & Recall & F$_1$ \\
        \midrule
        \textsc{REPaL} & 78.63 & \textbf{74.29} & \textbf{71.71} \\
        \quad -\,\texttt{implicit} & \textbf{80.60} & 71.15 & 71.40\\
        \quad\quad -\,\texttt{brief} & 80.00 & 71.08 & 71.56\\
        \bottomrule
    \end{tabular}
    \caption{Evaluation results w/o follow-up generation and conditioned on different initial positive generation templates. 30p30n training instances are gathered across all settings.}
    \label{tab:affect_pos_templates_fewrel_30p60n}
    \vspace{-8pt}
\end{table}

Table~\ref{tab:prompt templates for initial seed construction with LLM_brief}, Table~\ref{tab:prompt templates for initial seed construction with LLM_medium}, and Table~\ref{tab:prompt templates for initial seed construction with LLM_implicit} in Appendix~\ref{sec::appendix:Prompt Templates for Definition-Based Seed Construction with LLM} shows the adopted initial positive seed construction templates. 
Note that no ICL is adopted in \textsc{REPaL}'s generation step.
Analysis in Appendix~\ref{sec::appendix:Prompt Templates for Definition-Based Seed Construction with LLM} yields the conclusion that the generated instance's patterns generally follow the prompt instructions well, covering instances with brief, medium-length, and implicit patterns respectively.

The quantitative results with different combinations of positive instance generation prompts are shown in Table~\ref{tab:affect_pos_templates_fewrel_30p60n}.
We can see that results are generally robust against different prompt combinations. 
Leveraging all the \texttt{medium}, \texttt{brief}, and \texttt{implicit} prompts yields the best recall and slightly better overall F$_1$.
In our main experiments, all three prompts are adopted to enhance diversity of generated relation patterns. 

\subsection{Effect of More Iterations}\label{sec::results and analysis:effect of more iterations}

\begin{table}[ht]
    \small
    \centering
    \begin{tabular}{lccc}
        \toprule 
        \multirow{2}{*}{Iteration} & \multicolumn{3}{c}{DefOn-FewRel} \\
        \cmidrule(l){2-4}
        & Precision & Recall & F$_1$ \\
        \midrule
        1 & 73.59 & 75.90 & 70.03 \\
        2 & 78.86 & 77.28 & 74.61 \\
        3 & 80.61 & 74.91 & 74.57 \\
        4 & 78.85 & 76.93 & 74.96 \\
        \bottomrule
    \end{tabular}
    \caption{Results on DefOn-FewRel with different iterations based on \textsc{REPaL} w GPT-4o. Iteration 1 refers to the round of initial instance generation, and Iteration 2 refers to the first feedback-driven instance generation which is taken for the main experiment.}
    \label{tab:effect_of_more_iterations}
    \vspace{-12pt}
\end{table}

We further run \textsc{REPaL} with more iterations on the DefOn-FewRel dataset with results shown on Table~\ref{tab:effect_of_more_iterations}. 
The results indicates that more iterations can further improve the performance, but it exhibits a dynamic trade-off between precision and recall, a common challenge of learning with limited supervision, especially given our assumption of unknown negative relation space: 
As the model learns to recognize more true positive instances through the extended pattern coverage brought by synthesized samples, it may also include more noise (false positives). 
When the model corrects the false positives in the next iteration, it may become more conservative and lower the recall. 
Since conducting more rounds of iterative refinement incurs more costs, we leave more comprehensive explorations (e.g., performance change by more iterations, impact of LLM’s long-context capabilities) for future work.

\subsection{Error Analysis}\label{sec::appendix:error analysis}

\begin{table*}[!ht]
    \centering
    \resizebox{0.95\textwidth}{!}{
        \begin{tabular}{m{5cm}|m{8cm}|m{16cm}}
            \Xhline{1pt} 
            \multicolumn{1}{c|}{\textbf{Example Target Relation}} & \multicolumn{1}{c|}{\textbf{Majority False Positive Predicted Relations}} & \multicolumn{1}{c}{\textbf{Example Instance of False Positive Predicted Relations}} \\
            \Xhline{1pt} 
            P40: <ENT1> was/is the child (not stepchild) of <ENT0> & \tabincell{m{8cm}}{
            1. P26 \textcolor{red}{(129)}:	<ENT1> was/is the married spouse (husband, wife, partner, etc.) of <ENT0> \\ 
            2. P3373 \textcolor{red}{(183)}:	<ENT1> and <ENT0> had/have at least one common parent (<ENT1> is the sibling, brother, sister, etc. including half-sibling of <ENT0>) } & 
            \tabincell{m{16cm}}{
            1. Daughter of <ENT1> Sancho IV </ENT1> and of <ENT0> María de Molina </ENT0> , Infanta Beatrice was born in Toro . \textcolor{blue}{(Gold: P26 | Pred: P40 | Pos Prob: 0.861)} \\ 
            2. Sofia Coppola was born in New York City , New York , the youngest child and only daughter of set decorator / artist <ENT1> Eleanor Coppola </ENT1> ( née Neil ) and director <ENT0> Francis Ford Coppola </ENT0> . \textcolor{blue}{(Gold: P26 | Pred: P40 | Pos Prob: 0.665)}\\ 
            3. <ENT1> Ruby Aldridge </ENT1> is the daughter of former Playboy playmate Laura Lyons and artist and graphic designer Alan Aldridge , and younger sister of fashion model <ENT0> Lily Aldridge </ENT0> . \textcolor{blue}{(Gold: P3373 | Pred: P40 | Pos Prob: 0.885)} \\ 
            4.  Amongst his regular visitors were his younger brothers <ENT0> Jyotirindranath Tagore </ENT0> ( 1849–1925 ) and Rabindranath Tagore ( 1861–1941 ) , the Nobel Prize – winning poet , and his sister <ENT1> Swarnakumari Devi </ENT1> . \textcolor{blue}{(Gold: P3373 | Pred: P40 | Pos Prob: 0.945)}
            } \\
            \hline 
            P410: <ENT1> was/is the military rank achieved by or associated with <ENT0> (a person or a position) & 
            \tabincell{m{8cm}}{P241 \textcolor{red}{(210)}:	<ENT1> was/is the military branch to which <ENT0> (a military unit, award, office, or person) belonged/belongs} & 
            \tabincell{m{16cm}}{
            1. In November 1966 , retired <ENT1> USMC </ENT1> Major <ENT0> Donald Keyhoe </ENT0> and Richard H. Hall , both of NICAP , briefed the panel . \textcolor{blue}{(Gold: P241 | Pred: P410 | Pos Prob: 0.906)} \\ 
            2. <ENT0> Ricardo Sanchez </ENT0> ( born 1953 ) is a retired <ENT1> United States Army </ENT1> lieutenant general .  \textcolor{blue}{(Gold: P241 | Pred: P410 | Pos Prob: 0.768)}} \\
            \Xhline{1pt} 
        \end{tabular}
    }
    \caption{Error analysis of the predictions made by SLM-based RE model. The contents in red denote the number of false positive predictions for a specific relation. The contents in blue denote the prediction details made by SLM-based RE model. \textit{Gold} refers to the gold relation label of an instance. \textit{Pred} refers to the predictions made by our model. \textit{Pos Porb} means the predicted probability of the instance following the target relation. }
    \label{tab:error analysis}
    \vspace{-14pt}
\end{table*}

Table~\ref{tab:error analysis} shows the major source of false positive predictions of the final tuned SLM RE model, it can be seen that the majority of false positive predictions are concentrated on a few similar negative relations. 
As our proposed task setting assumes the unknown negative relation space, one challenge appears to be learning the positive relation against the unknown and infinitely many negative relations. 
To address such challenge, our model derive targeted negative relations based on the feedback of model inference. 
Based on the results in Table~\ref{tab:error analysis}, we can see that the challenge is not fully eliminated which serves as a promising future research direction. 
Another feature seen from the false positive predicted instances is that some typical false positive instances actually express the target relation in addition to its gold relation.
However, the target relation is not expressed by the tagged entity mention pair.
This may indicate that better RE architectures which well model the position awareness of target entity pairs can be adopted for improving the overall performance.

\vspace{-6pt}

\section{Related Work}

\paragraph{Zero-Shot Relation Extraction}
Our work is related to zero-shot RE \citep{levy2017zero}. 
Majority zero-shot RE approaches mainly leverage clustering, label-verbalization, or Siamese-based architectures \citep{rahimi2023improving, chen2021zs, chia2022relationprompt, li2023revisiting} which seek for the instance-instance similarity or the similarity between the relation instances and the unseen relations' information. 
\citet{chen2021zs} utilize relation descriptions for zero-shot RE but their approach still relies on seen data to align relation descriptions with instances in a supervised manner.
\citet{li-etal-2023-semi} adopt the relation descriptions but only for verifying synthesized data with the instance-level seeds. 
LLM-based RE works focus on designing prompting strategies or LLM alignment to tackle zero-shot RE \citep{li2023revisiting, zhang2023aligning, wei2023zero, wadhwa-etal-2023-revisiting}.
Our work is distinguished from pre-LLM zero-shot RE works as they heavily rely on the supervision from massive seen data and the complete negative relation space. 
And the majority do not focus on relation definitions. 
Our work is different from LLM-based zero-shot RE works as we emphasize both the rich relation definitions for data synthesis and synergy between SLM and LLM.

\paragraph{Definition-Driven Text Mining} 
BERTNet \citep{hao-etal-2023-bertnet} applies definitions for distilling entities from LM parametric knowledge. 
Label definitions/descriptions have also been proven to be powerful in text classification \citep{gao-etal-2023-benefits}.
In zero-shot RE, several PLM-based works have utilized relation definitions \citep{chen2021zs, zhao2023re} but they mainly focus on computing instance-definition similarities.
In our work, LLM is used to distill patterns and extend or rectify the learning of SLM based on definitions.

\section{Conclusion}
In this work, we have introduced a new zero-shot RE task where only relation definitions instead of seen-unseen relation instances are provided. 
Correspondingly, we have proposed \textsc{REPaL} which leverages LLMs and unlabeled corpora to generate relation instances and iteratively self-improves the generation pattern coverage while rectifying the bias by automatically acquiring and reflecting over sampled feedback from multiple sources.
Quantitative experiments and qualitative analysis on our two modified datasets show the effectiveness and robustness of our framework as well as our large-margin advance over most baselines. 
Exploratory experiments show that generating more data in a single-turn conversation does not yield proportionally larger pattern coverage. 
We also proposed a derive-definition-then-generate approach which achieves much better performance than just utilizing few-shot instances. 
This gives insights into low-resource RE works to capture the complete relation semantics to avoid partial coverage by few-shot instances. 

\clearpage

\section*{Limitations}
In this work, we mainly experimented on GPT for data synthesis as their instruction following performance is competent so that we do not need to introduce in-context learning in most of the time. 
Therefore, one follow-up work is to explore other LLMs to see their generation capability compared to the GPT series.
Besides, new RE datasets tailored for our definition only zero-shot RE still can be created as it still lacks large scale yet high quality datasets.
Thirdly, prompt engineering and hyperparameter search are not conducted. 
For the sake of better performance in downstream tasks, future works could compensate this.

\section*{Ethics Statements}
Since our goal is to solve sentence-level RE tasks where the text contexts are sufficient to derive the relation, factualness of the relation triples is not a strict requirement or a vital factor for the training instances.
Therefore, in generative data synthesis, we do not further verify the factualness of the generation results and we simply count on the LLMs. 
Therefore follow-up works could explore this and other related approaches should also be careful if they want to adapt our work to downstream tasks necessitating factualness such as factual question answering.

\section*{Acknowledgements}
Research was supported in part by the Institute for Geospatial Understanding through an Integrative Discovery Environment (I-GUIDE) by NSF under Award No. 2118329, US DARPA INCAS Program No. HR0011-21-C0165 and BRIES Program No. HR0011-24-3-0325, National Science Foundation IIS-19-56151, and the Molecule Maker Lab Institute: An AI Research Institutes program supported by NSF under Award No. 2019897. Any opinions, findings, and conclusions or recommendations expressed herein are those of the authors and do not necessarily represent the views, either expressed or implied, of DARPA or the U.S. Government.

\bibliography{custom}

\clearpage
\appendix
\section{Detailed Discussions on Definition Only Zero-Shot Relation Extraction Setup}\label{sec::appendix:Detailed Discussions on Definition Only Zero-Shot Relation Extraction Setup}
\subsection{Task Setup and Evaluation Process}
Suppose the RE test set contains $R$ relations and each relation $r$ has $N_r$ instances. 
Then the evaluation will be conducted in $R$ iterations and the final score will be calculated by averaging over the individual scores from the total $R$ iterations.

For each test iteration, we take one test relation $r$ as the target relation (gold positive relation) with $N_r$ gold positive instances. 
All the other test relations will be treated as negative relations and all their associated instances are gold negative instances. 
So each iteration is in the format of binary classification, targeting at the set of gold positive instances and gold negative instances. 
Additionally, we assume the negative relation space is unknown, which means that for each test iteration, we only know the relation definition of the gold positive relation $r$, and we do not know any information about the negative test relations. 
Such cross-validation evaluation is designed for robustness as the results are averaged over all different test relations.
For the evaluation complexity, if we assume each test relation has balanced $N$ instances, the complexity is proportional to $N \cdot R^2$.

\subsection{Distinctions with Traditional Zero-Shot RE}
Traditional zero-shot relation extraction models are trained on a large set of seen relations with corresponding labeled instances. 
During evaluation, the trained model will deal with a set of unseen relations with corresponding relation instances. 
The trained model will have access to the information of all unseen relations (at the same time) in the format such as relation label names, or relation descriptions/definitions, or other more fine-grained information (e.g., extended possible entity type lists). 
Then the final model is required to conduct multi-way classification over all unseen relations.

For our newly defined Definition Only Zero-Shot Relation Extraction task:
\begin{itemize}
    \item We do not rely on any seen relation or any labeled relation instance.
    \item We only assume a clear and complete relation definition for each target positive relation and an unlabeled corpus.
    \item We assume unknown negative relation space which means for each test iteration, in addition to the single positive target test relation, all the other test relations are treated as negative relations and we do not know any information about how many negative relations are and what the negative relations are.
    \item The evaluation process is completely different from traditional zero-shot RE as described above.
\end{itemize}

\subsection{Motivations and Practical Values}
We design our Definition Only Zero-Shot Relation Extraction in order to accommodate more realistic and more challenging applications as the fast developing LMs, especially LLMs, are enabling the design of such advanced systems. 
Here we will emphasize the motivations and values of the setup of our Definition Only Zero-Shot Relation Extraction task.

First, the assumption that the definition of the target unseen relation is given instead of assuming and using annotated data of other seen relations accommodates various applications scenarios. 
When people deal with domain specific problems, the definitions of interested relations are normally clear and explicit. 
For instance, an expert in the geographic information system (GIS) domain might want to model textual patterns which describe two geospatial entities “touches” with each other (their interiors do not intersect and only their boundaries intersect). 
They already have well defined terminologies and associated definitions for such relations but the annotations are expensive. 
Another example lies in the question answering task where one can derive the “<ENT0> is youngest birth child of <ENT1>” definition if they are interested in extracting instances (text contexts and tuples) for the question “Who is the youngest child of Person A?”. 
Additionally, when crowdsource workers are annotating RE samples, they are often provided with the relation definitions to guide their annotation work. 
Therefore it’s realistic to assume that a clear and explicit relation definition is given. 
From the above cases, it can also be seen that the potential relations of interest are infinite which emerge with different problems in different domains. 
But annotating in-domain samples are expensive and time consuming. 
It’s hence also meaningful to tackle RE from the source. 
Namely, focusing on definitions which capture the complete relation semantics compared to other types of starting seeds and leveraging the domain adaptation power and the constrained generation power of LLMs to alleviate the annotation scarcity issue.

Second, the assumption of unknown negative relation space is to mimic the real world setting where the number of interested relations is extremely small compared to the number of negative relations between entities. For example, one may be interested in one Wikidata property relation but the number of total Wikidata properties is massive. 
In such cases, approximately, we barely know anything about the negative relation space. 
But to train a good relation extractor that can distinguish interested target relations against all the other negative relations in the corpus, we will need some method designs to deal with such unknown negative relation space. 
Our assumption of unknown negative relation space can also be considered as a more challenging version of ``None'' or ``No Relation'' relation labels for traditional multi-classification RE datasets.

Third, the assumption of a large unlabeled corpus is natural as the unlabeled corpus usually come together with specific domains.
Still taking the geospatial RE as an example, it’s relatively easy to acquire documents that mention about geospatial entities and potentially their mutual relations. 
Some experts from GIS could also be able to provide such corpus. 
Since the corpus do not need to be labeled with relations, it’s much more convenient and efficient to get such unlabeled corpus set up instead of gathering domain-specific annotations. 
Besides, our framework does not solely rely on the unlabeled corpus as we have designed follow-up positive and negative instance generation processes to rectify the bias and extend the relation pattern coverage which synthesizes follow-up positive and negative instances. 
LLMs will conduct self-reflection on the given relation definition, synthesis dialogue history, and the sampled inference results on the unlabeled corpora. 
Among those sampled inference results, LLM will judge whether the prediction is correct or wrong. 
If the sampled inference results contain correct predictions, LLM can analyze the instances which convey the information on the patterns learnt by SLM. 
If the sampled inference results contain wrong predictions, LLM can summarize their relations and generate follow-up similar negative relations and corresponding instances to correct the learning of SLM. 
Even if the feedback does not contain useful information, LLM can still generate more positive relation patterns based on the previously generated instances. 
LLM can also generate near-miss negative relations simply based on the definition of the positive target relation. 
With LLMs becoming more powerful in inference and having longer context window sizes, the performance gain of follow-up feedback driven generation process can be further improved.

Finally, our setting of binary NLI is versatile and has great potential to adapt to multi-way classification and multi-label classification settings. 
Based on the task setup and evaluation process, we can see there will be one binary relation classifier for each target relation. 
This is versatile because if we stick to the multi-classification setting based on data synthesis approach, we would need to re-train our multi-way classifier on synthesized data if there is an additional new relation coming in. 
Furthermore, the setup of $R$ binary classifiers accommodates the scenarios where there are relations entailed by other relations or the contexts indicate two possible relations which are not allowed by multi-way classification setup. 
If we have multiple interested relations and the number of such relations are large, there are various methods to reduce the cost of adapting binary classifiers for multi-class/label problems. 
One method is to use some rules (e.g., mismatched entity types) or some coarse-grained NLI methods to first filter impossible relation candidates to reduce the candidate relation space and then apply our trained relation classifiers.

\section{Prompt Templates Used in Definition-Based Seed Construction with LLM}\label{sec::appendix:Prompt Templates for Definition-Based Seed Construction with LLM}
\begingroup
\begin{table*}[ht]
    \centering
    \small
    \begin{tabular}{p{0.92\linewidth}}
        \toprule
        \underline{\textbf{\textsc{Prompt Template For Initial Positive Instance Generation (brief)}}} \\
        \vspace{-2mm}
        A binary relation between entity placeholders <ENT0> and <ENT1> is defined by ``\hl{\{relation\_definition\}}''. Under sentence-level relation extraction setting, generate \hl{\{number\_of\_examples\}} examples (numbered from 1 to \hl{\{number\_of\_examples\}}) expressing the same relation, where <ENT0> is replaced with actual entity mention and is prefixed with tag <ENT0> and suffixed with tag <\/ENT0> , <ENT1> is replaced with actual entity mention and is prefixed with tag <ENT1> and suffixed with <\/ENT1> . Do not overfit the pattern of the definition. Try as many different relation patterns or relation expressions as possible.\\
        \bottomrule
    \end{tabular}
    \caption{
        Prompt templates (\texttt{brief}) used in Definition-Based Seed Construction with LLM (Sec.~\ref{sec::methodology:def-based initial seed construction}). Words highlighted denote the placeholders for filling in contents indicated by their surface names. 
    }
    \label{tab:prompt templates for initial seed construction with LLM_brief}
\end{table*}
\endgroup

\begingroup
\begin{table*}[ht]
    \centering
    \small
    \begin{tabular}{p{0.92\linewidth}}
        \toprule
        \underline{\textbf{\textsc{Prompt Template For Initial Positive Instance Generation (medium)}}} \\
        \vspace{-2mm}
        A binary relation between entity placeholders <ENT0> and <ENT1> is defined by ``\hl{\{relation\_definition\}}''. Under sentence-level relation extraction setting, generate \hl{\{number\_of\_examples\}} examples (numbered from 1 to \hl{\{number\_of\_examples\}}) expressing the same relation, where <ENT0> is replaced with actual entity mention and is prefixed with tag <ENT0> and suffixed with tag <\/ENT0> , <ENT1> is replaced with actual entity mention and is prefixed with tag <ENT1> and suffixed with <\/ENT1> . Other content requirements:\\
        \\
        1. Do not overfit the pattern of definition. Try as many different relation patterns or relation expressions as possible.\\
        2. Generate rich and informative related contexts before and after each entity.\\
        \bottomrule
    \end{tabular}
    \caption{
        Prompt templates (\texttt{medium}) used in Definition-Based Seed Construction with LLM (Sec.~\ref{sec::methodology:def-based initial seed construction}). Words highlighted denote the placeholders for filling in contents indicated by their surface names. 
    }
    \label{tab:prompt templates for initial seed construction with LLM_medium}
\end{table*}
\endgroup

\begingroup
\begin{table*}[ht]
    \centering
    \small
    \begin{tabular}{p{0.92\linewidth}}
        \toprule
        \underline{\textbf{\textsc{Prompt Template For Initial Positive Instance Generation (implicit)}}} \\
        \vspace{-2mm}
        A binary relation between entity placeholders <ENT0> and <ENT1> is defined by ``\hl{\{relation\_definition\}}''. Under sentence-level relation extraction setting, generate \hl{\{number\_of\_examples\}} examples (numbered from 1 to \hl{\{number\_of\_examples\}}) expressing the same relation, where <ENT0> is replaced with actual entity mention and is prefixed with tag <ENT0> and suffixed with tag <\/ENT0> , <ENT1> is replaced with actual entity mention and is prefixed with tag <ENT1> and suffixed with <\/ENT1> . Other content requirements:\\
        \\
        1. Do not overfit the pattern of definition. Try as many different relation patterns or relation expressions as possible.\\
        2. Generate rich and informative related contexts before and after each entity.\\
        3. The relation patterns or relation expressions should be implicit or complicated\\
        \bottomrule
    \end{tabular}
    \caption{
        Prompt templates (\texttt{implicit}) used in Definition-Based Seed Construction with LLM (Sec.~\ref{sec::methodology:def-based initial seed construction}). Words highlighted denote the placeholders for filling in contents indicated by their surface names. 
    }
    \label{tab:prompt templates for initial seed construction with LLM_implicit}
\end{table*}
\endgroup
\begin{table*}[!ht]
    \centering
    \resizebox{\textwidth}{!}{
        \begin{tabular}{m{2cm}|m{18cm}}
            \Xhline{1pt} 
            \multicolumn{1}{c|}{\textbf{Prompt Template Name}} & \multicolumn{1}{c}{\textbf{Example Generated Instances (Relation=P241)}} \\
            \Xhline{1pt} 
            \multicolumn{1}{c|}{\texttt{brief}} & \tabincell{m{18cm}}{
            1. During World War II, <ENT0>The 101st Airborne Division</ENT0> proudly served under <ENT1>The United States Army</ENT1>, where they played a crucial role in several pivotal battles. \\ 
            2. The prestigious <ENT0>Victoria Cross</ENT0> is awarded by <ENT1>The British Army</ENT1> for acts of extraordinary valor in the presence of the enemy. \\ 
            3. <ENT0>General Douglas MacArthur</ENT0> was a prominent figure in <ENT1>The United States Army</ENT1>, noted for his leadership in the Pacific Theater during World War II.
            } \\
            \hline 
            \multicolumn{1}{c|}{\texttt{medium}} & \tabincell{m{18cm}}{
            1. <ENT0>The U.S. Navy SEALs</ENT0>, an elite special operations force, is a notable unit that belongs to the <ENT1>United States Navy</ENT1>, which plays a crucial role in maritime security and warfare. \\ 
            2. During his illustrious career, <ENT0>General Dwight D. Eisenhower</ENT0> served in the <ENT1>United States Army</ENT1>, playing a key role during World War II and later becoming the 34th President of the United States. \\ 
            3. The prestigious <ENT0>Distinguished Flying Cross</ENT0> is an honor awarded to personnel of the <ENT1>United States Air Force</ENT1> for acts of heroism or extraordinary achievement during aerial flight.
            } \\
            \hline 
            \multicolumn{1}{c|}{\texttt{implicit}} & \tabincell{m{18cm}}{
            1. During the freezing winter of 1944, the decisive Battle of the Bulge tested the mettle of many military entities, among them <ENT0>the 101st Airborne Division</ENT0>. Engaged in ferocious combat, the valor of these troops was on full display under the aegis of the <ENT1>United States Army</ENT1>. \\ 
            2. Last summer, the grand ceremony at the Capitol honored various noteworthy figures, including <ENT0>General Dwight D. Eisenhower</ENT0>, whose illustrious career and leadership were long-standing pillars of the <ENT1>United States Army</ENT1>. \\ 
            3. On Veterans Day, numerous speeches commemorated those it was instituted to serve, like <ENT0>Sergeant John Doe</ENT0>, a brave soul who once operated under the proud tradition and command structure of the <ENT1>Marine Corps</ENT1>.
            } \\
            \Xhline{1pt} 
        \end{tabular}
    }
    \caption{Prompt templates used in Definition-Based Seed Construction with LLM (Sec.~\ref{sec::methodology:def-based initial seed construction}). Example generated instances are based on the relation P241: <ENT1> was/is the military branch to which <ENT0> (a military unit, award, office, or person) belonged/belongs.}
    \label{tab:case study for initial seed construction with LLM}
    
\end{table*}

Table~\ref{tab:prompt templates for initial seed construction with LLM_brief}, Table~\ref{tab:prompt templates for initial seed construction with LLM_medium}, and Table~\ref{tab:prompt templates for initial seed construction with LLM_implicit} contains the three prompt templates used for generating initial positive seeds using LLMs. 

Example instances generated by corresponding prompts are shown in Table~\ref{tab:case study for initial seed construction with LLM}. 
Our goal of designing such prompts is to cover all the patterns for target relations.
From the generated example instances, it is evident that the pattern complexity (or, more simply, the sentence length) shows considerable differences, particularly between the \texttt{implicit} prompt and the other two prompts.
The pattern complexity (or more na\"{i}vely, the sentence length) well follows the instructions conveyed by each type of prompt and well represents the prompt name, \texttt{brief}, \texttt{medium}, and \texttt{implicit} correspondingly.

\section{Prompt Templates Used in Feedback-Driven Instance improvement and Bias Rectification}\label{sec::appendix:Prompt Templates Used in Feedback-Driven Instance improvement and Bias Rectification}
\begingroup
\begin{table*}[ht]
    \centering
    \small
    \begin{tabular}{p{0.92\linewidth}}
        \toprule
        \underline{\textbf{\textsc{Prompt Template For Follow-Up Positive Instance Generation}}} \\
        \vspace{-2mm}
        Sampled examples which are predicted as positive by my relation extraction model are:\\
        \\
        \hl{\{feedback\_examples\}}  \\
        \\
        Based on these predicted examples and your previously generated examples, generate \hl{\{number\_of\_examples\}} additional examples (numbered from 1 to \hl{\{number\_of\_examples\}}) expressing the same pre-defined relation: ``\hl{\{relation\_definition\}}''. Other requirements are:\\
        \\
        1. Identify what relation patterns have been learnt by my model and what relation patterns have been covered by your previously generated examples. Your newly generated examples should have different and diverse relation patterns. \\
        2. Identify model's bias from the sampled predicted examples which do not express the correct relation definition and your newly generated examples should try to mitigate the bias.\\
        3. If the sampled predicted examples are uninformative, focus on the dialogue history, especially examples that were previously generated, to generate new examples with different and more diverse patterns.
        \\
        \bottomrule
    \end{tabular}
    \caption{
        Prompt template used for follow-up positive instance generation in Feedback-Driven Instance Improvement and Bias Rectification (Sec.~\ref{sec::methodology:feedback-driven instance improve and bias rectify}). Words highlighted denote the placeholder for filling in contents indicated by their surface names. 
    }
    \label{tab:prompt templates for Feedback-Driven Instance Improvement and Bias Rectification_Follow-Up Positive Instance Generation}
\end{table*}
\endgroup

\begingroup
\begin{table*}[ht]
    \centering
    \small
    \begin{tabular}{p{0.92\linewidth}}
        \toprule
        \underline{\textbf{\textsc{Prompt Template For Follow-Up Negative Relation Definition Generation}}} \\
        \vspace{-2mm}
        A binary relation between entity placeholders <ENT0> and <ENT1> is defined by: ``\hl{\{positive\_relation\_definition\}}''. In relation examples or relation instances, <ENT0> is replaced with actual entity mention and is prefixed with tag <ENT0> and suffixed with tag </ENT0> , <ENT1> is replaced with actual entity mention and is prefixed with tag <ENT1> and suffixed with </ENT1> .\\
        Typical examples predicted as positive by current relation extraction model are:\\
        \\
        \hl{\{feedback\_examples\}} \\
        \\ 
        Based on the positive relation definition and the typical predicted examples, generate \hl{\{number\_of\_negative\_relations\}} negative binary relation definitions (numbered from 1 to  \hl{\{number\_of\_negative\_relations\}}) in the same format as the above positive relation definition (including entity placeholders and entity type constraints). Other requirements are:\\
        \\
        1. Identify false positive predictions from the typical predicted examples and your generated negative relations should teach model to mitigate such bias.\\
        2. After addressing the previous requirement or if there is no false positive prediction, consider generating near-miss negative relations.
        \\
        \bottomrule
    \end{tabular}
    \caption{
        Prompt template used for follow-up negative relation definition generation (without previously generated negative relation definitions) in Feedback-Driven Instance Improvement and Bias Rectification (Sec.~\ref{sec::methodology:feedback-driven instance improve and bias rectify}). Words highlighted denote the placeholder for filling in contents indicated by their surface names. 
    }
    \label{tab:prompt templates for Feedback-Driven Instance Improvement and Bias Rectification_Follow-Up Negative Relation Definition Generation wo Prev Gen Neg Rels}
\end{table*}
\endgroup

\begingroup
\begin{table*}[ht]
    \centering
    \small
    \begin{tabular}{p{0.92\linewidth}}
        \toprule
        \underline{\textbf{\textsc{Prompt Template For Follow-Up Negative Relation Definition Generation}}} \\
        \vspace{-2mm}
        A binary relation between entity placeholders <ENT0> and <ENT1> is defined by: ``\hl{\{positive\_relation\_definition\}}'' (in relation examples, <ENT0> is replaced with actual entity mention and is prefixed with tag <ENT0> and suffixed with tag </ENT0> , <ENT1> is replaced with actual entity mention and is prefixed with tag <ENT1> and suffixed with </ENT1> .).\\
        Typical examples predicted as positive by current relation extraction model are:\\
        \\
        \hl{\{feedback\_examples\}} \\
        \\ 
        Existing generated negative relation definitions are:\\
        \\
        \hl{\{previously\_generated\_negative\_relation\_definitions\}} \\
        \\
        Based on the positive relation definition, the typical predicted examples and the existing generated negative relation definitions, generate \hl{\{number\_of\_negative\_relations\}}  additional negative binary relation definitions (numbered from 1 to  \hl{\{number\_of\_negative\_relations\}}) in the same format as the above positive relation definition (including entity placeholders and entity type constraints). Other requirements are:\\
        \\
        1. Identify false positive predictions from the typical predicted examples and your generated negative relations should teach model to mitigate such bias.\\
        2. After addressing the previous requirement or if there is no false positive prediction, consider generating near-miss negative relations. \\
        3. Your generated negative relation definitions should not be the same as existing negative relation definitions.
        \\
        \bottomrule
    \end{tabular}
    \caption{
        Prompt template used for follow-up negative relation definition generation (with previously generated negative relation definitions) in Feedback-Driven Instance Improvement and Bias Rectification (Sec.~\ref{sec::methodology:feedback-driven instance improve and bias rectify}). Words highlighted denote the placeholder for filling in contents indicated by their surface names. 
    }
    \label{tab:prompt templates for Feedback-Driven Instance Improvement and Bias Rectification_Follow-Up Negative Relation Definition Generation w Prev Gen Neg Rels}
\end{table*}
\endgroup

Table~\ref{tab:prompt templates for Feedback-Driven Instance Improvement and Bias Rectification_Follow-Up Positive Instance Generation}, Table~\ref{tab:prompt templates for Feedback-Driven Instance Improvement and Bias Rectification_Follow-Up Negative Relation Definition Generation wo Prev Gen Neg Rels}, and Table~\ref{tab:prompt templates for Feedback-Driven Instance Improvement and Bias Rectification_Follow-Up Negative Relation Definition Generation w Prev Gen Neg Rels}
contain prompt templates for our feedback-driven follow-up positive instance generation and negative relation definition generation respectively.
Note that Table~\ref{tab:prompt templates for Feedback-Driven Instance Improvement and Bias Rectification_Follow-Up Negative Relation Definition Generation w Prev Gen Neg Rels}, which assumes there are previously generated negative relation definitions, is required for \textsc{REPaL} iterations following the first feedback-driven generation iteration.
After obtaining the negative relation definitions, we simply leverage the \texttt{medium} instance generation template in Table~\ref{tab:prompt templates for initial seed construction with LLM_medium} to generate negative relation instances. 
We take this template as our purpose of negative instance generation is to rectify the existing bias instead of pursuing complete negative relation pattern coverage. 
Furthermore, Table~\ref{tab:affect_pos_templates_fewrel_30p60n} demonstrates that the performance difference between the usages of different templates is minor.

\section{Details of Constructed DefOn-FewRel and DefOn-Wiki-ZSL Datasets}\label{sec::appendix:Details on Constructed DefOn-FewRel and DefOn-Wiki-ZSL}
\begin{table*}[!ht]
    \centering
    \resizebox{\textwidth}{!}{
        \begin{tabular}{m{3cm}|m{12cm}|m{3cm}}
            \Xhline{1pt} 
            \multicolumn{1}{c|}{\textbf{Original Dataset}} & \multicolumn{1}{c|}{\textbf{Relation Label : Definition}} & \multicolumn{1}{c}{\textbf{Frequency}} \\
            \Xhline{1pt} 
            \multicolumn{1}{c|}{FewRel} & P106: <ENT1> was/is the occupation of <ENT0> (a person) & \multicolumn{1}{c}{700} \\
            \hline 
            \multicolumn{1}{c|}{FewRel} & P1344: <ENT0> (a person or organization) was/is a participant of <ENT1> (an event) & \multicolumn{1}{c}{700} \\
            \hline 
            \multicolumn{1}{c|}{FewRel} & P136: <ENT1> was/is the genre or the field of work of <ENT0> (a creative work or an artist) & \multicolumn{1}{c}{700} \\
            \hline 
            \multicolumn{1}{c|}{FewRel} & P1411: <ENT1> was/is the award nomination received by <ENT0> (a person, organisation, or creative work) & \multicolumn{1}{c}{698} \\
            \hline 
            \multicolumn{1}{c|}{FewRel} & P241: <ENT1> was/is the military branch to which <ENT0> (a military unit, award, office, or person) belonged/belongs & \multicolumn{1}{c}{700} \\
            \hline 
            \multicolumn{1}{c|}{FewRel} & P26: <ENT1> was/is the married spouse (husband, wife, partner, etc.) of <ENT0> & \multicolumn{1}{c}{700} \\
            \hline 
            \multicolumn{1}{c|}{FewRel} & P276: <ENT1> was/is the location of <ENT0> (an object, structure or event) & \multicolumn{1}{c}{700} \\
            \hline 
            \multicolumn{1}{c|}{FewRel} & P3373: <ENT1> and <ENT0> had/have at least one common parent (<ENT1> is the sibling, brother, sister, etc. including half-sibling of <ENT0>) & \multicolumn{1}{c}{700} \\
            \hline 
            \multicolumn{1}{c|}{FewRel} & P40: <ENT1> was/is the child (not stepchild) of <ENT0> & \multicolumn{1}{c}{700} \\
            \hline 
            \multicolumn{1}{c|}{FewRel} & P400: <ENT1> was/is the platform or platform version for which <ENT0> (a work or a software product) was/is developed or released & \multicolumn{1}{c}{700} \\
            \hline 
            \multicolumn{1}{c|}{FewRel} & P410: <ENT1> was/is the military rank achieved by or associated with <ENT0> (a person or a position) & \multicolumn{1}{c}{700} \\
            \hline 
            \multicolumn{1}{c|}{FewRel} & P57: <ENT1> was/is the director(s) of <ENT0> (a film, TV-series, stageplay, video game or similar) & \multicolumn{1}{c}{700} \\
            \hline 
            \multicolumn{1}{c|}{FewRel} & P84: <ENT1> was/is the architect or architectural firm responsible for designing <ENT0> (a building) & \multicolumn{1}{c}{700} \\
            \hline 
            \multicolumn{1}{c|}{FewRel} & P974: <ENT1> was/is the watercourse that flowed/flows into <ENT0> (a watercourse) & \multicolumn{1}{c}{700} \\
            \Xhline{1pt} 
        \end{tabular}
    }
    \caption{Example DefOn-FewRel relation labels, definitions, and corresponding instance frequencies.}
    \label{tab:defon fewrel relation distributions and definitions}
\end{table*}
Table~\ref{tab:defon fewrel relation distributions and definitions} shows the example relation labels and constructed definitions.
Please refer to our Github repository for detailed relations and definitions. 
To get quality evaluation samples, we conduct test data cleaning with the requirements as: (1) The two entity mentions should not overlap; (2) The entity mentions should not be pronouns such as \textit{I}, \textit{he}, and \textit{she}.
Note these two requirements only give negligible impact on the number of relation instances. 

Furthermore, we clean the unlabeled corpora before down-sampling by requiring that selected unlabeled samples should not be repeated.
Namely, for any two unlabeled samples, the sentence, the head entity mention and the tail entity mention can not all be the same.

\section{Effect of More Initial Seeds}
\begin{figure*}[!th]
    \centering
    \includegraphics[width=0.9\linewidth]{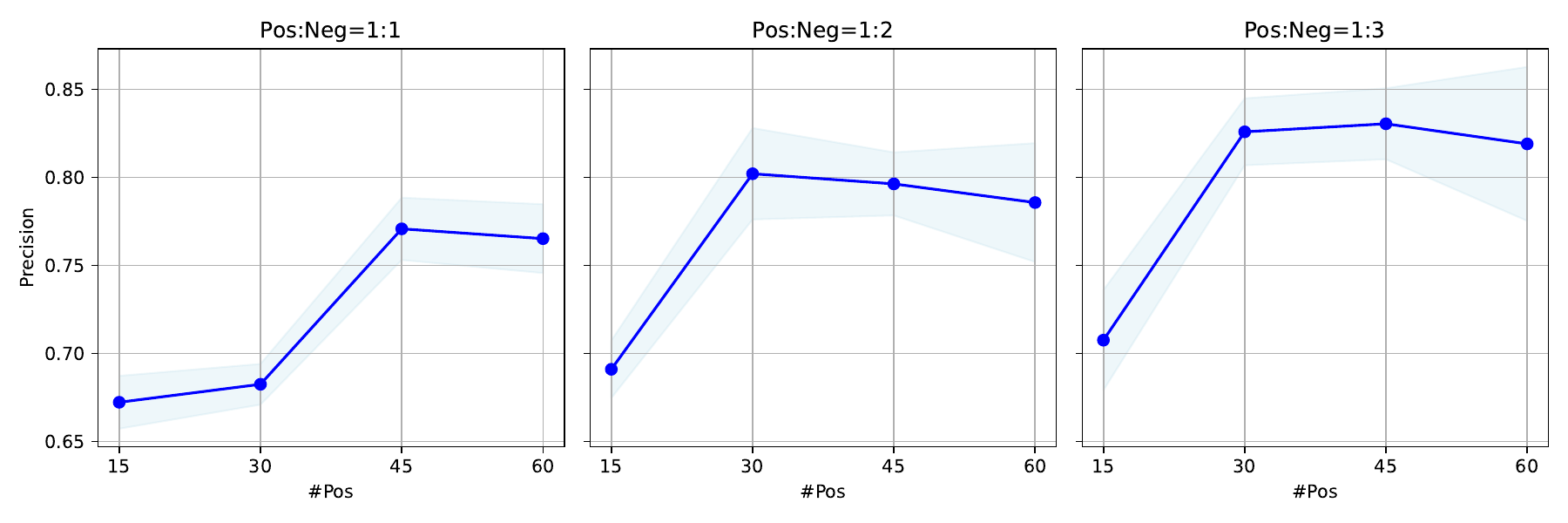}
    \caption{Precision (\%) scores for different setups on the number and ratio of training instances.}
    \label{fig:num_ratio_init_seeds_prec}
\end{figure*}

\begin{figure*}[!th]
    \centering
    \includegraphics[width=0.9\linewidth]{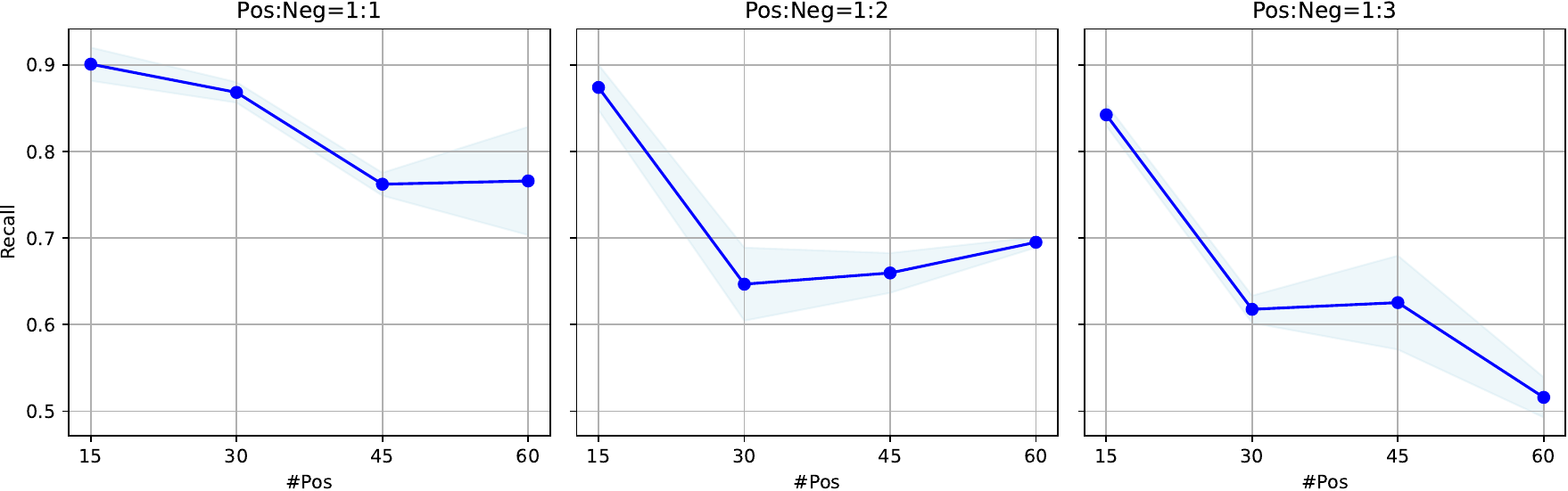}
    \caption{Recall (\%) scores for different setups on the number and ratio of training instances.}
    \label{fig:num_ratio_init_seeds_recall}
\end{figure*}

The precision (\%) scores and recall (\%) scores discussed in Sec.~\ref{sec::Background:effect of more initial seeds} are shown in Fig.~\ref{fig:num_ratio_init_seeds_prec} and Fig.~\ref{fig:num_ratio_init_seeds_recall} respectively.

\section{Implementation Details}\label{sec::appendix:implementation details}

\subsection{Baselines}
In consideration of OpenAI API calling expense, the GPT-3.5 baseline results are from the evaluation over down-sampled test sets (30 down-sampled test instances for each relation).
The prompt template used for GPT-3.5 baseline is shown by Table~\ref{tab:gpt35 infer prompt template}. 
The prompt templates used for \textsc{RE as QA} on our \textit{Definition Only Zero-Shot RE } setting and on traditional multi-class classification setting are shown by Table~\ref{tab:RE as QA infer prompt template_defon} and Table~\ref{tab:RE as QA infer prompt template_multiway} respectively. 
For inference, we use the temperature as 0.
For all the other baselines, we adopted their original hyperparameters reported in corresponding publications. 
\begingroup
\begin{table*}[ht]
    \centering
    \small
    \begin{tabular}{p{0.92\linewidth}}
        \toprule
        \underline{\textbf{\textsc{Prompt Template For Inference with GPT-3.5 Baseline}}} \\
        \vspace{-2mm}
        A binary relation between entity placeholders <ENT0> and <ENT1> is defined by ``\hl{\{relation\_definition\}}''. In following relation instances, <ENT0> will be replaced with actual entity mention and prefixed with tag <ENT0> and suffixed with tag </ENT0> , <ENT1> will be replaced with actual entity mention and prefixed with tag <ENT1> and suffixed with </ENT1> . Now given an instance: ``\hl{\{instance\_sentence\_with\_entities\_enclosed\_by\_tags\}}'', choose one option to answer: is the relation between two entities in the instance the same as the defined positive relation?\\
        Option 1: Yes\\
        Option 2: No\\
        \\
        Answer:
        \\
        \bottomrule
    \end{tabular}
    \caption{
        Prompt templates used for GPT-3.5 baseline in main experiments. Words highlighted denote the placeholders for filling in contents indicated by their surface names. 
    }
    \label{tab:gpt35 infer prompt template}
\end{table*}
\endgroup

\begingroup
\begin{table*}[ht]
    \centering
    \small
    \begin{tabular}{p{0.92\linewidth}}
        \toprule
        \underline{\textbf{\textsc{Prompt Template For RE as QA Inference on Definition Only Zero-Shot RE}}} \\
        \vspace{-2mm}
        A binary relation between entity placeholders <ENT0> and <ENT1> is defined by ``\hl{\{relation\_definition\}}''. In the following relation instances, <ENT0> will be replaced with actual entity mention and prefixed with tag <ENT0> and suffixed with tag </ENT0> , <ENT1> will be replaced with actual entity mention and prefixed with tag <ENT1> and suffixed with </ENT1> .\\
        Given an instance, choose one option to answer: based on the new instance, do the entity mention enclosed by <ENT0> and </ENT0> and the entity mention enclosed by <ENT1> and </ENT1> follow the above positive relation definition? The chosen option should come from:\\
        Option 1: Yes\\
        Option 2: No\\
        \\
        Now answer:\\
        Instance: \hl{\{instance\_sentence\_with\_entities\_enclosed\_by\_tags\}} \\
        Question: is the following statement true based on the instance: \hl{\{relation\_definition\_filled\_with\_instance\_entities\}}\\
        Answer: 
        \\
        \bottomrule
    \end{tabular}
    \caption{
        Prompt templates used for \textsc{RE as QA} baseline in main experiments following \textit{Definition Only Zero-Shot Relation Extraction} setting. Words highlighted denote the placeholders for filling in contents indicated by their surface names. 
    }
    \label{tab:RE as QA infer prompt template_defon}
\end{table*}
\endgroup

\begingroup
\begin{table*}[ht]
    \centering
    \small
    \begin{tabular}{p{0.92\linewidth}}
        \toprule
        \underline{\textbf{\textsc{Prompt Template For RE as QA Inference on Multi-Class Zero-Shot RE}}} \\
        \vspace{-2mm}
        Determine which option can be inferred from the given sentence.\\
        \\
        Sentence: \hl{\{instance\_sentence\}} \\
        \\
        Options:\\
        A. \hl{\{definition\_of\_relation\_1\_filled\_with\_instance\_entities\}} \\
        B. \hl{\{definition\_of\_relation\_2\_filled\_with\_instance\_entities\}}\\
        ...omitted for clarity... \\
        N. \hl{\{definition\_of\_relation\_14\_filled\_with\_instance\_entities\}} \\
        \\
        Respond with one letter from ``A''-``N''.
        \\
        \bottomrule
    \end{tabular}
    \caption{
        Prompt templates used for \textsc{RE as QA} baseline in main experiments following traditional multi-class classification zero-shot RE setting. Words highlighted denote the placeholders for filling in contents indicated by their surface names. 
        This template is adapted from QA4RE \citep{zhang2023aligning}.
        This prompt template assumes the number of all relations is 14 for demonstration purpose. 
    }
    \label{tab:RE as QA infer prompt template_multiway}
\end{table*}
\endgroup

\subsection{\textsc{REPaL}}
We use \texttt{gpt-4o-2024-05-13} for GPT-4o and \texttt{gpt-4o-mini-2024-07-18} for GPT-4o mini. 
We have max tokens set to 4,096, presence penalty set to 0 and temperature set to 0.6.  
The SLM checkpoint used is \textsc{roberta-large-mnli}. 

For main experiments and analysis with \textsc{REPaL}, in addition to constructing the train set, we also construct a dev set with same number of positive instances and negative instances as the train set to automatically select the SLM model checkpoint for test. 
The dev positive instances are obtained by running the corresponding positive training instance generation with exactly the same prompt input and same setup for another time.
The dev negative instances are obtained by random sampling from the unlabeled corpus.

For the step of \textit{Definition-Based Seed Construction with LLM and Unlabeled Corpus}, we follow the templates in Table~\ref{tab:prompt templates for initial seed construction with LLM_brief}, Table~\ref{tab:prompt templates for initial seed construction with LLM_medium}, and Table~\ref{tab:prompt templates for initial seed construction with LLM_implicit}. 
For the results reported in Table~\ref{tab:main_eval_results}, we choose the setting of 15 initial positive seeds and 15 initial negative seeds based on the analysis in Sec.~\ref{sec::Background:effect of more initial seeds}. 
As we have three types of prompts for positive seeds generation, each prompt will contribute to generating 5 initial positive seeds. 
For the step of \textit{Pattern Learning with RE-Specialized SLM}, we train the SLM with 12 epochs using AdamW optimizer \citep{loshchilov2017decoupled} with learning rate equal to 3e-5 and batch size equal to 64. 

For the step of \textit{Feedback-Driven Instance Improvement and Bias Rectification}\footnote{Note that for this step, we adopt GPT-4o mini as a  parser to obtain the list of generated negative relation definitions or generated instances from the output string of synthesis LLM. This can help separate the generated definitions/instances from the additional output analysis.}, we first leverage the trained SLM to conduct inference on the unlabeled corpus after which each unlabeled instance will be associated with a score as the probability of being positive. 
Then, for follow-up positive instance generation, we conduct random sampling from all instance with score higher than 0.85 as we want the sampled feedback instances to reflect the model's learning outcome for the target positive relation. 
For follow-up negative relation definition generation, we conduct random sampling from all instances with score higher than 0.50 as we want to see both the confident predictions and less confident predictions to identify the existing bias. 
Note that continuing from the initial positive seed generation, there will be three threads of dialogue history for follow-up positive instance generation corresponding to three types of prompts in Table~\ref{tab:prompt templates for initial seed construction with LLM_brief}, Table~\ref{tab:prompt templates for initial seed construction with LLM_medium}, and Table~\ref{tab:prompt templates for initial seed construction with LLM_implicit} respectively.
So for each thread of dialogue, we fill in different groups of sampled feedback instances to maximize the feedback coverage. 
For both the follow-up positive instance generation and the follow-up negative instance generation, the number of feedback instances for each prompt input is set to 10.
For follow-up negative relation definition generation, we set the number of generated negative relation definitions to be 5 and the number of total follow-up negative instances to be 15. 
After obtaining the feedback-driven follow-up instances, we repeat the SLM training with all the accumulated training instances and all the other hyperparameters the same as our previous SLM training step.

\section{LLM-Based Relation Definition Derivation}

\begin{table*}[!ht]
    \centering
    \resizebox{\textwidth}{!}{
        \begin{tabular}{m{5cm}|m{15cm}}
            \Xhline{1pt} 
            \multicolumn{1}{c|}{\textbf{Prompt Name}} & \multicolumn{1}{c}{\textbf{Prompt Template}} \\
            \Xhline{1pt} 
            \multicolumn{1}{c|}{\texttt{Few-Shot Definition Derivation}} & Given a list of relation instances/examples of a binary relation defined between two entities <ENT0> and <ENT1>, derive the relation definition in a single sentence. Note that in relation instances/examples, actual entity mention for <ENT0> is prefixed with tag <ENT0> and suffixed with tag </ENT0>, and actual entity mention for <ENT1> is prefixed with tag <ENT1> and suffixed with </ENT1> . Your derived relation definition should use entity placeholders <ENT0> and <ENT1> to refer to the two entities and the relation definition should try to contain entity type constraints. Example relation definitions are:\textbackslash{}n\textbackslash{}n1. <ENT1> is the league in which <ENT0> (team or player) plays or has played in.\textbackslash{}n\textbackslash{}n2. <ENT1> is the organization or person responsible for publishing <ENT0> (books, periodicals, printed music, podcasts, games or software).\textbackslash{}n\textbackslash{}n3. <ENT1> is the city, where <ENT0> (an organization)'s headquarters is or has been situated.\textbackslash{}n\textbackslash{}nThe list of relation instances/examples is:\textbackslash{}n\textbackslash{}n\textcolor{blue}{\$\{Few-Shot Instances for One Relation\}}\textbackslash{}n\textbackslash{}n \\
            \hline
            \multicolumn{1}{c|}{\texttt{Train Instance Generation}} & A binary relation between entity placeholders <ENT0> and <ENT1> is defined by ``\textcolor{blue}{\$\{Derived Relation Definition\}}''. Under sentence-level relation extraction setting, generate additional \textcolor{blue}{\$\{Number Of Additional Examples to Generate\}} examples (numbered from 1 to \textcolor{blue}{\$\{Number Of Additional Examples to Generate\}} expressing the same relation, where <ENT0> is replaced with actual entity mention and is prefixed with tag <ENT0> and suffixed with tag </ENT0> , <ENT1> is replaced with actual entity mention and is prefixed with tag <ENT1> and suffixed with </ENT1> . \textcolor{blue}{\$\{Gold Few-Shot Examples for ICL\}} Do not overfit the pattern of the definition. Try as many different relation patterns or relation expressions as possible. \\
            \Xhline{1pt} 
        \end{tabular}
    }
    \caption{Prompt template used in deriving original relation definitions given few-shot relation instances and generating new relation instances based on the derived relation definition and gold few-shot instances (Sec.~\ref{sec::Background:Derive Relation Definition from Few-Shot Instances}). Words in blue denote the placeholder for filling in contents indicated by their surface names. }
    \label{tab:fewshot2def prompts}
    
\end{table*}

Our adopted prompt template for deriving relation definitions based on few-shot instances in Sec~\ref{sec::Background:Derive Relation Definition from Few-Shot Instances} is shown in Table~\ref{tab:fewshot2def prompts}. 
Note that we leveraged a fixed 3 relation definition demonstrations for in-context learning across all relations so that the LLM can give the relation definition in our desired format for automatic parsing. 
After getting the relation definition, we use the prompt template in Table~\ref{tab:fewshot2def prompts}
to generate 15 instances for each derived relation. 
Note that the instance generation prompt is basically the same as \texttt{brief} prompt in Table~\ref{tab:prompt templates for initial seed construction with LLM_brief} except that it integrates the gold few-shot instances as in-context learning demonstrations.

\subsection{Relation Definitions Derived by LLM From Few-Shot Instances}\label{sec::appendix:Relation Definitions Derived by LLM From Few-Shot Instances}
\begin{table*}[!ht]
    \centering
    \resizebox{0.9\textwidth}{!}{
        \begin{tabular}{m{3cm}|m{18cm}|m{3cm}}
            \Xhline{1pt} 
            \multicolumn{1}{c|}{\textbf{Gold Definition}} & \multicolumn{1}{c|}{\textbf{Gold Few-Shot Instances For Derivation}} & \multicolumn{1}{c}{\textbf{Derived Definition}} \\
            \Xhline{1pt} 
            \tabincell{m{3cm}}{<ENT1> was/is the occupation of <ENT0> (a person)} & \tabincell{m{18cm}}{1. <ENT0>Pierre Maudru</ENT0> ( 1892\textbackslash{}u20131992 ) was a French <ENT1>screenwriter</ENT1> . Goble p.189 He also directed three films .\\ 2. WWF Hall of Famer Bob Backlund and Extreme Championship Wrestling <ENT1>manager</ENT1> <ENT0>Bill Alfonso</ENT0> also made surprise appearances during the event . \\ 3. In May 2010 , Paratici moved from Sampdoria to Juventus , along with Director General Giuseppe Marotta and <ENT1>Manager</ENT1> <ENT0>Luigi Delneri</ENT0> . \\ 4. <ENT0>Else Reval</ENT0> ( 14 June 1893 \textbackslash{}u2013 25 January 1978 ) was a German <ENT1>film actress</ENT1> . Giesen p.210} & \tabincell{m{3cm}}{<ENT1> is the profession in which <ENT0> (a person) works or has worked.} \\
            \hline
            \tabincell{m{3cm}}{<ENT0> (a person or organization) was/is a participant of <ENT1> (an event)} & \tabincell{m{18cm}}{1. He only saw limited action in <ENT1>Euro 2000</ENT1> as cover for left - back <ENT0>Arthur Numan</ENT0> . \\ 2. <ENT0>Francesco Cameli</ENT0> was a sailor from Italy , who represented his country at the <ENT1>1928 Summer Olympics</ENT1> in Amsterdam , Netherlands . \\ 3. <ENT0>Giannin Andreossi</ENT0> ( born July 2 , 1902 , date of death unknown ) was a Swiss ice hockey player who competed in the <ENT1>1928 Winter Olympics</ENT1> . \\ 4. <ENT0>Ren\textbackslash{}u00e9 Sch\textbackslash{}u00f6fisch</ENT0> ( born February 3 , 1962 ) is a German speed skater who competed for East Germany in the <ENT1>1984 Winter Olympics</ENT1> .} & \tabincell{m{3cm}}{<ENT1> is the major international sports competition in which <ENT0> (an athlete) has competed.} \\
            \hline
            \tabincell{m{3cm}}{<ENT1> was/is the genre or the field of work of <ENT0> (a creative work or an artist)} & \tabincell{m{18cm}}{1. Another version , dating from c. 1616 , was given in c. 1790 to <ENT0>Joshua Reynolds</ENT0> by the Duke of Leeds in exchange for a Reynolds self - <ENT1>portrait</ENT1> .\\2. Teixeira is a former member of indie rock bands Ik Mux and Boris Ex - Machina , as well as the <ENT1>hip hop</ENT1> group <ENT0>Da Weasel</ENT0> and industrial metal band Bizarra Locomotiva .\\3. Beautiful Stories for Ugly ChildrenMUSHROOMHEAD To Release ' Beautiful Stories For Ugly Children ' In September is the seventh studio album by <ENT1>industrial metal</ENT1> band <ENT0>Mushroomhead</ENT0> .\\4. Wales is portrayed in the 1976 <ENT1>western film</ENT1> `` <ENT0>The Outlaw Josey Wales</ENT0> '' by actor and director Clint Eastwood .} & \tabincell{m{3cm}}{<ENT1> is the genre or type of art (music, painting, film) associated with <ENT0> (an artist, band, or cultural artifact).} \\
            \hline
            \tabincell{m{3cm}}{<ENT1> was/is the award nomination received by <ENT0> (a person, organisation, or creative work)} & \tabincell{m{18cm}}{1. On January 24 , 2012 , he was nominated for an <ENT1>Academy Award for Best Adapted Screenplay</ENT1> for the movie `` <ENT0>Moneyball</ENT0> '' .\\ 2. `` <ENT0>The Great Santini</ENT0> '' received two Academy Award nominations : <ENT1>Best Actor in a Leading Role</ENT1> ( Duvall ) and Best Actor in a Supporting Role ( O'Keefe ) .\\ 3. `` <ENT0>Born This Way</ENT0> '' ( 2011 ) , Gaga 's second studio album , accrued three nominations at the 54th Annual Grammy Awards , including her third consecutive nomination for <ENT1>Album of the Year</ENT1> .\\ 4. As a producer , he has been nominated for <ENT1>Best Picture</ENT1> for three other films : `` Raging Bull '' , `` <ENT0>The Right Stuff</ENT0> '' , and `` Goodfellas '' .} & \tabincell{m{3cm}}{<ENT1> is the award category for which <ENT0> (films, albums, or individuals associated with entertainment productions) has been nominated.} \\
            \hline
            \tabincell{m{3cm}}{<ENT1> was/is the military branch to which <ENT0> (a military unit, award, office, or person) belonged/belongs} & \tabincell{m{18cm}}{1. General Sir ( William ) <ENT0>Henry Mackinnon</ENT0> , ( 15 December 1852 \textbackslash{}u2013 17 March 1929 ) was a <ENT1>British Army</ENT1> General during World War I.\\2. Lieutenant - Colonel <ENT0>Gordon Graham Donaldson</ENT0> was a senior officer in the <ENT1>British Army</ENT1> who died as a result of illness contracted during the disastrous Walcheren Campaign in 1809 .\\3. <ENT0>Raphael Semmes</ENT0> was an officer in the <ENT1>United States Navy</ENT1> from 1826 to 1860 and the Confederate States Navy from 1860 to 1865 .\\4. <ENT0>Isaac Townsend</ENT0> ( `` c. '' 1685 \textbackslash{}u2013 21 November 1765 ) was an admiral in the <ENT1>British Royal Navy</ENT1> and a Member of Parliament .} & \tabincell{m{3cm}}{<ENT1> is the military organization (such as an army or navy) with which <ENT0> (an individual, specified by their role or rank) has served or been associated.} \\
            \hline
            \tabincell{m{3cm}}{<ENT1> was/is the married spouse (husband, wife, partner, etc.) of <ENT0>} & \tabincell{m{18cm}}{1. The film is about <ENT0>Carolyn Cassady</ENT0> 's recollection of life with husband <ENT1>Neal Cassady</ENT1> and Jack Kerouac , and her concern that the truth about these men is being lost in their mythos .\\2. Maximilian married Duchess Helene in Bavaria , daughter of <ENT1>Duke Maximilian Joseph in Bavaria</ENT1> and his wife <ENT0>Princess Ludovika of Bavaria</ENT0> , on 24 August 1858 at Possenhofen Castle .\\3. In 1916 his younger daughter , <ENT0>Nadejda</ENT0> ( `` Nada '' ) married <ENT1>Prince George of Battenberg</ENT1> , older son of Prince Louis by Queen Victoria 's granddaughter , Princess Victoria of Hesse - Darmstadt .\\4. The fourth and youngest son of King <ENT1>John II of France</ENT1> and his wife , <ENT0>Bonne of Luxembourg</ENT0> , Philip was the founder of the Burgundian branch of the House of Valois .} & \tabincell{m{3cm}}{<ENT1> is the spouse or partner of <ENT0> (an individual), indicating a marital, romantic, or partnership connection between the two entities.} \\
            \hline
            \tabincell{m{3cm}}{ <ENT1> was/is the location of <ENT0> (an object, structure or event)} & \tabincell{m{18cm}}{1. At the <ENT0>2014 Winter Olympics</ENT0> , Hudec won the bronze medal in the super - G at <ENT1>Rosa Khutor</ENT1> .\\2. On the night of 22 January 1942 during the <ENT0>Battle of the Points</ENT0> , Japanese troops of the 16th Division attempted a landing on the west coast of southern <ENT1>Bataan</ENT1> .\\3. Since the Netherlands did boycott the Moscow Olympic Games Brasser represented his National Olympic Committee at the <ENT0>1980 Summer Olympics</ENT0> in <ENT1>Tallinn</ENT1> , USSR under the Dutch NOC flag .\\4. The bridge Norrbro stretches past the Riksdag on <ENT0>Helgeandsholmen</ENT0> and further south to <ENT1>Stockholm Old Town</ENT1> and the Royal Palace .} & \tabincell{m{3cm}}{<ENT1> is the location or venue where <ENT0> (an event such as sports competitions, battles, or significant historical or cultural events) took place or was hosted.} \\
            \hline
            \tabincell{m{3cm}}{<ENT1> and <ENT0> had/have at least one common parent (<ENT1> is the sibling, brother, sister, etc. including half-sibling of <ENT0>)} & \tabincell{m{18cm}}{1. Together they had three sons : Antonio , <ENT1>Arturo</ENT1> , and <ENT0>Alejandro</ENT0> .\\2. Portuguese and Spanish conquerors made use of these weapons , including Vasco da Gama and his sons <ENT1>Crist\textbackslash{}u00f3v\textbackslash{}u00e3o da Gama</ENT1> and the younger brother <ENT0>Est\textbackslash{}u00eav\textbackslash{}u00e3o da Gama</ENT0> .\\3. <ENT1>Arjuna</ENT1> was the fourth one to fall after Draupadi , <ENT0>Sahadeva</ENT0> and Nakula .\\4. His nephews , Andr\textbackslash{}u00e9 , <ENT0>Jordan</ENT0> and <ENT1>Rahim</ENT1> , also played the sport professionally .} & \tabincell{m{3cm}}{<ENT1> is the sibling, specifically the brother, of <ENT0>.} \\
            \Xhline{1pt} 
        \end{tabular}
    }
    \caption{Comparison between gold relation definitions and few-shot (4-shot) derived relation definitions (random seed=1).}
    \label{tab:comparison between gold relation definitions and few-shot derived relation definitions}
    
\end{table*}

\begin{table*}[!ht]
    \centering
    \resizebox{0.9\textwidth}{!}{
        \begin{tabular}{m{3cm}|m{18cm}|m{3cm}}
            \Xhline{1pt} 
            \multicolumn{1}{c|}{\textbf{Gold Definition}} & \multicolumn{1}{c|}{\textbf{Gold Few-Shot Instances For Derivation}} & \multicolumn{1}{c}{\textbf{Derived Definition}} \\
            \Xhline{1pt}
            \tabincell{m{3cm}}{<ENT1> was/is the child (not stepchild) of <ENT0>} & \tabincell{m{18cm}}{1. He was the son of Flemish painter <ENT1>Jan Massys , Matsys , or Metsys</ENT1> and the grandson and namesake of <ENT0>Quentin Massys or Metsys</ENT0> .\\2. She married <ENT1>Lu Jing</ENT1> , who was born to <ENT0>Lu Kang</ENT0> and another daughter of Zhang Cheng ; both Sun He 's daughter and Lu Jing therefore were Zhang Cheng 's maternal grandchildren .\\3. She is the wife of Bollywood actor , <ENT0>Jackie Shroff</ENT0> and mother of <ENT1>Tiger Shroff</ENT1> and Krishna Shroff .\\4. His uncle was polymath <ENT0>Lionel Penrose</ENT0> , whose children include mathematician <ENT1>Oliver Penrose</ENT1> , polymath Sir Roger Penrose , chess grandmaster Jonathan Penrose , and geneticist Shirley Hodgson .} & \tabincell{m{3cm}}{<ENT1> is a direct family member (such as a son, grandson, wife, or mother) of <ENT0>, specified by their familial relationship.} \\
            \hline
            \tabincell{m{3cm}}{<ENT1> was/is the platform or platform version for which <ENT0> (a work or a software product) was/is developed or released} & \tabincell{m{18cm}}{1. The <ENT1>NES</ENT1> version of <ENT0>Shadowgate</ENT0> also carries the distinction of being one of the few NES games to be available in a Swedish language version .\\2. In case of incidents <ENT0>Plumbr</ENT0> provides its users with information on problem severity , problem 's root cause location in source code or <ENT1>JVM</ENT1> configuration and lists steps needed to take to remediate the problem .\\3. In 2013 , `` <ENT0>Mega Man Xtreme</ENT0> '' was made available on the Virtual Console of Japan 's Nintendo eShop for the <ENT1>Nintendo 3DS</ENT1> .\\4. Prior to <ENT0>Windows 2000</ENT0> , Windows NT ( and thus PE ) supported the MIPS , Alpha , and <ENT1>PowerPC</ENT1> ISAs .} & \tabincell{m{3cm}}{<ENT1> is the platform, console, or environment for which <ENT0> (software applications, games, or operating systems) is designed or available.} \\
            \hline
            \tabincell{m{3cm}}{<ENT1> was/is the military rank achieved by or associated with <ENT0> (a person or a position)} & \tabincell{m{18cm}}{1. The son of Robert Langton Douglas , he was a half - brother to <ENT1>Marshal of the Royal Air Force</ENT1> <ENT0>William Sholto Douglas , 1st Baron Douglas of Kirtleside</ENT0> .\\2. <ENT0>Dwight Edward Aultman</ENT0> , <ENT1>Brigadier General</ENT1> , United States Army . \\3. He then served in the 27th U - boat Flotilla along with `` <ENT1>Korvettenkapit\u00e4n</ENT1> '' <ENT0>Erich Topp</ENT0> .\\4. <ENT0>Axel Schimpf</ENT0> ( born 1 October 1952 ) is a retired `` <ENT1>Vizeadmiral</ENT1> '' ( vice admiral ) of the German Navy .} & \tabincell{m{3cm}}{<ENT1> is the military rank of <ENT0> (a military personnel).} \\
            \hline
            \tabincell{m{3cm}}{<ENT1> was/is the director(s) of <ENT0> (a film, TV-series, stageplay, video game or similar)} & \tabincell{m{18cm}}{1. Cummins 's photographs have been used extensively in cinema and TV documentaries including <ENT1>Grant Gee</ENT1> 's  <ENT0>Joy Division</ENT0>  and John Dower 's .\\2. In 2014 , Zhang starred in <ENT1>Tsui Hark</ENT1> 's wuxia film `` <ENT0>The Taking of Tiger Mountain</ENT0> '' .\\3. Starting her career in 2005 , she acted in the Malayalam film `` <ENT0>Boyy Friennd</ENT0> '' directed by <ENT1>Vinayan</ENT1> .\\4. Kaif had her first success in Bollywood when she appeared opposite Salman Khan in <ENT1>David Dhawan</ENT1> 's romantic comedy `` <ENT0>Maine Pyaar Kyun Kiya ?</ENT0> '' .} & \tabincell{m{3cm}}{<ENT1> is the director who directed the movie <ENT0>.} \\
            \hline
            \tabincell{m{3cm}}{<ENT1> was/is the architect or architectural firm responsible for designing <ENT0> (a building)} & \tabincell{m{18cm}}{1. The <ENT0>Hotel Attraction</ENT0> project by Catalan architect <ENT1>Antoni Gaud\textbackslash{}u00ed</ENT1> was built in 1908 in the parallel universe , whereas in our world it never went past initial planning .\\2. Tampere Cathedral ( Lars Sonck , 1900 ) , <ENT0>National Museum</ENT0> , Helsinki ( <ENT1>Herman Gesellius , Armas Lindgren and Eliel Saarinen</ENT1> , 1902 ) .\\3. Its designer was <ENT1>George Gilbert Scott</ENT1> , <ENT0>Busbridge Church</ENT0> \textbackslash{}2013 Church of England Official gateway to the church .\\4. He served a seven - year apprenticeship with <ENT1>Sir Charles Barry</ENT1> , the architect of the <ENT0>Houses of Parliament</ENT0> and Manchester Art Gallery .} & \tabincell{m{3cm}}{<ENT1> is the architect or group of architects who designed <ENT0> (a building or architectural project).} \\
            \hline
            \tabincell{m{3cm}}{<ENT1> was/is the watercourse that flowed/flows into <ENT0> (a watercourse)} & \tabincell{m{18cm}}{1. The <ENT1>Cerchez River</ENT1> is a tributary of the <ENT0>Ceair River</ENT0> in Romania .\\2. The lake flows into the <ENT1>River Mangfall</ENT1> , a tributary of the <ENT0>River Inn</ENT0> and thence the River Danube .\\3. The <ENT1>Veljul Mic River</ENT1> is a tributary of the <ENT0>Veljul Mare River</ENT0> in Romania\\4. A small part of the district along the eastern boundary drains into the east - flowing <ENT1>River Loud</ENT1> , a tributary of the <ENT0>Hodder</ENT0> .} & \tabincell{m{3cm}}{<ENT1> is a tributary of the <ENT0> (rivers or water bodies).} \\
            \hline
            \Xhline{1pt} 
        \end{tabular}
    }
    \caption{(Continued from Table~\ref{tab:comparison between gold relation definitions and few-shot derived relation definitions}) Comparison between gold relation definitions and few-shot (4-shot) derived relation definitions (random seed=1).}
    \label{tab:continued-comparison between gold relation definitions and few-shot derived relation definitions}
    
\end{table*}

Table~\ref{tab:comparison between gold relation definitions and few-shot derived relation definitions} and Table~\ref{tab:continued-comparison between gold relation definitions and few-shot derived relation definitions} show the LLM derived relation definitions based on the gold 4-shot instances. 
The table also contains the ground truth relation definitions for reference.
We can see that for most of the FewRel relations, LLM successfully recovers the gold relation definitions. 
The derived definitions also reveal that one major difficulty is to specify the entity type constraints as few-shot instances may only convey a partial set of entity types which misguides LLMs to deduce a partial entity type constraints in the derived relation definitions.

\subsection{Macro F1 Scores of Few-Shot Method against Definition-Based Method}\label{sec::appendix:Macro F1 Scores of Few-Shot Method against Definition-Based Method}
\begin{figure}[!ht]
    \centering
    \includegraphics[width=0.9\linewidth]{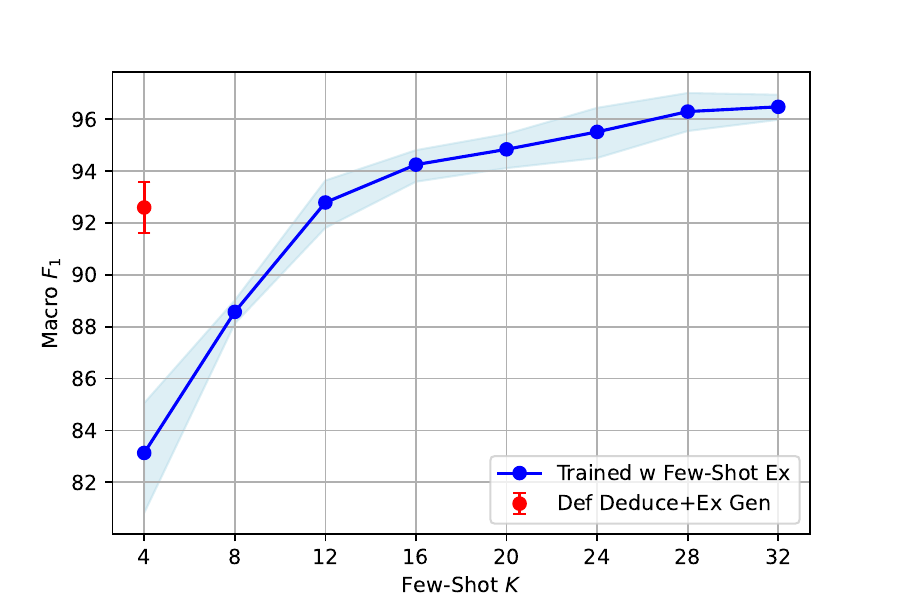}
    \caption{Macro F$_1$ (\%) scores of model trained with few-shot instances (\textit{Trained w Few-Shot Ex}) and model trained with instances from our relation definition derivation and instance generation approach (\textit{Def Deduce+Ex Gen}). The error bar/band denotes averaged value $\pm$ standard deviation.}
    \label{fig:def_derive_fewrel_macro_f1}
\end{figure}

The macro F$_1$ scores of the experiments conducted in Sec.~\ref{sec::Background:Derive Relation Definition from Few-Shot Instances} are shown in Table~\ref{fig:def_derive_fewrel_macro_f1}.
Since the DefOn-FewRel dataset is almost balanced, the micro F$_1$ and macro F$_1$ are close.
So we put macro F$_1$ here for reference.

\section{Fewshot Performance on DefOn-FewRel}\label{sec::appendix:Fewshot Performance on DefOn-FewRel}
\begin{table}[ht]
    \small
    \centering
    \begin{tabular}{lccc}
        \toprule 
        \multirow{2}{*}{ICL Demos} & \multicolumn{3}{c}{DefOn-FewRel} \\
        \cmidrule(l){2-4}
        & Precision & Recall & F$_1$ \\
        \midrule
        0p0n & 87.26 & 69.76 & 72.00 \\
        2p2n & 75.15 & 86.43 & 76.81 \\
        5p5n & 70.35 & 93.33 & 76.60 \\
        \bottomrule
    \end{tabular}
    \caption{Results of \textsc{RE as QA} on one split of DefOn-FewRel with increasing ICL demonstrations. 0p0n means zero-shot, while 2p2n means 2 sampled positive instance and 2 sampled gold negative instance as demonstration exemplars for each target relation. 5p5n means 5 sampled positive instance and 5 sampled gold negative instance as demonstration exemplars for each relation.}
    \label{tab:fewshot RE as QA inference}
    \vspace{-12pt}
\end{table}
We experiment inference with \textsc{RE as QA} on one split of DefOn-FewRel dataset by increasing the number of in-context learning exemplars. 
The results are shown in Table~\ref{tab:fewshot RE as QA inference}.
It shows that certain amount of in-context learning exemplars can lead to better performance while more exemplars do not necessarily result in better performance. 
It indicates that relation definitions are the most essential part of our \textit{Definition Only Zero-Shot RE} setting as definitions convey more complete relation semantics while exemplars could guide models to better comprehend the task and enhance the reasoning process.

Besides, our work mainly focuses on zero-shot RE which is a different setting from few-shot RE, and most existing zero-shot RE works do not include few-shot PLM/LLM methods as comparison baselines. 
Meanwhile, LLM-based inference is costly. 
\textsc{REPaL} could also replace its SLM with LLMs or introduce in-context learning strategies for better performance, but the increased computational cost and the reduced flexibility are against our low-resource positioning. 
Therefore, we stick with an efficient small encoder approach and leverage LLMs for zero-shot synthesis. 
The synthesized relation instances can be reused to train other relation extractors.

\section{Case Study}\label{sec::appendix:case study}

\begin{figure*}[!ht]
    \centering
    \includegraphics[width=0.8\linewidth]{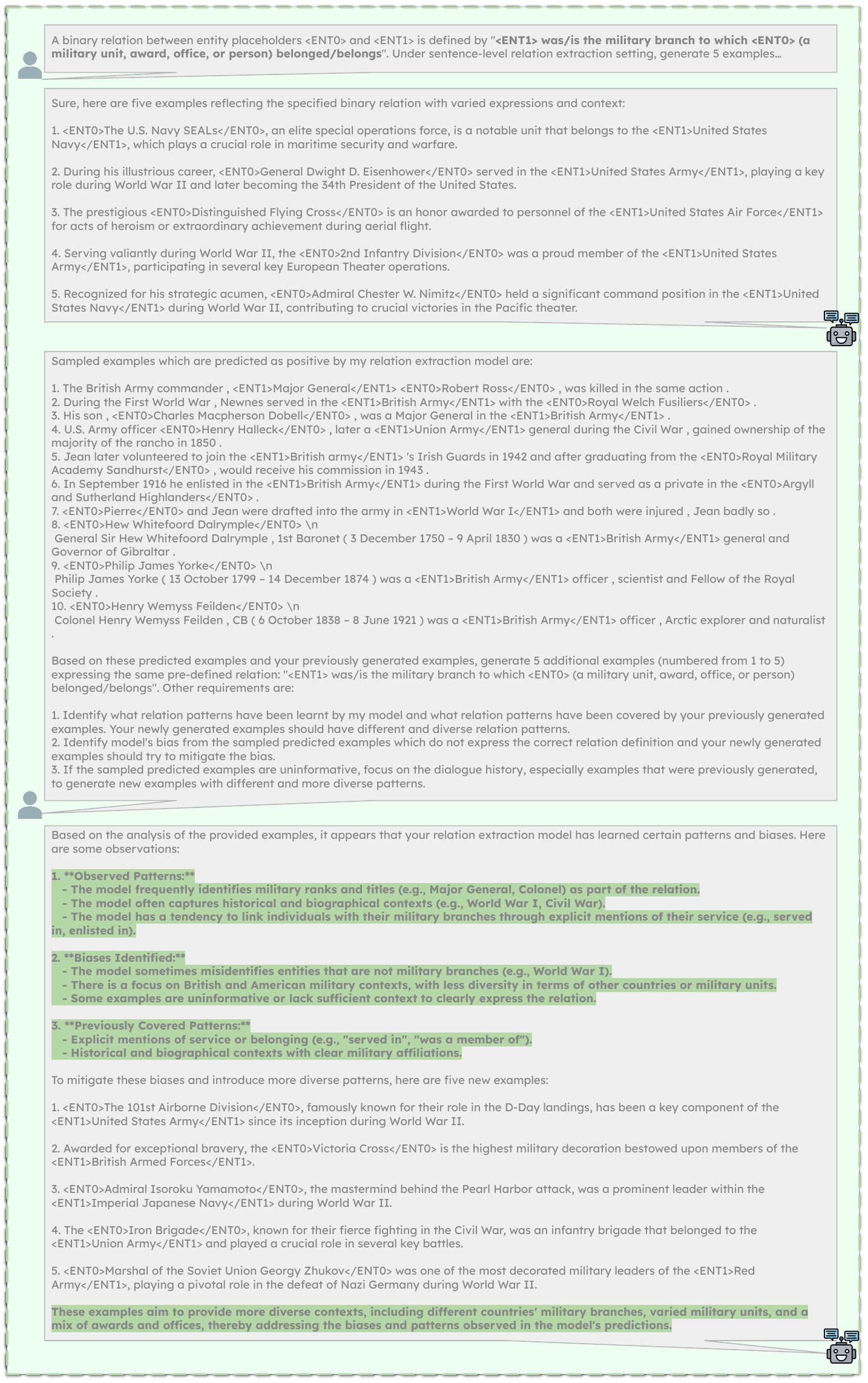}
    \caption{Example interaction dialogue which demonstrates the initial seed generation and feedback-driven follow-up positive instance generation. }
    \label{fig:case study for pos gen}
\end{figure*}

\begin{figure*}[!ht]
    \centering
    \includegraphics[width=0.8\linewidth]{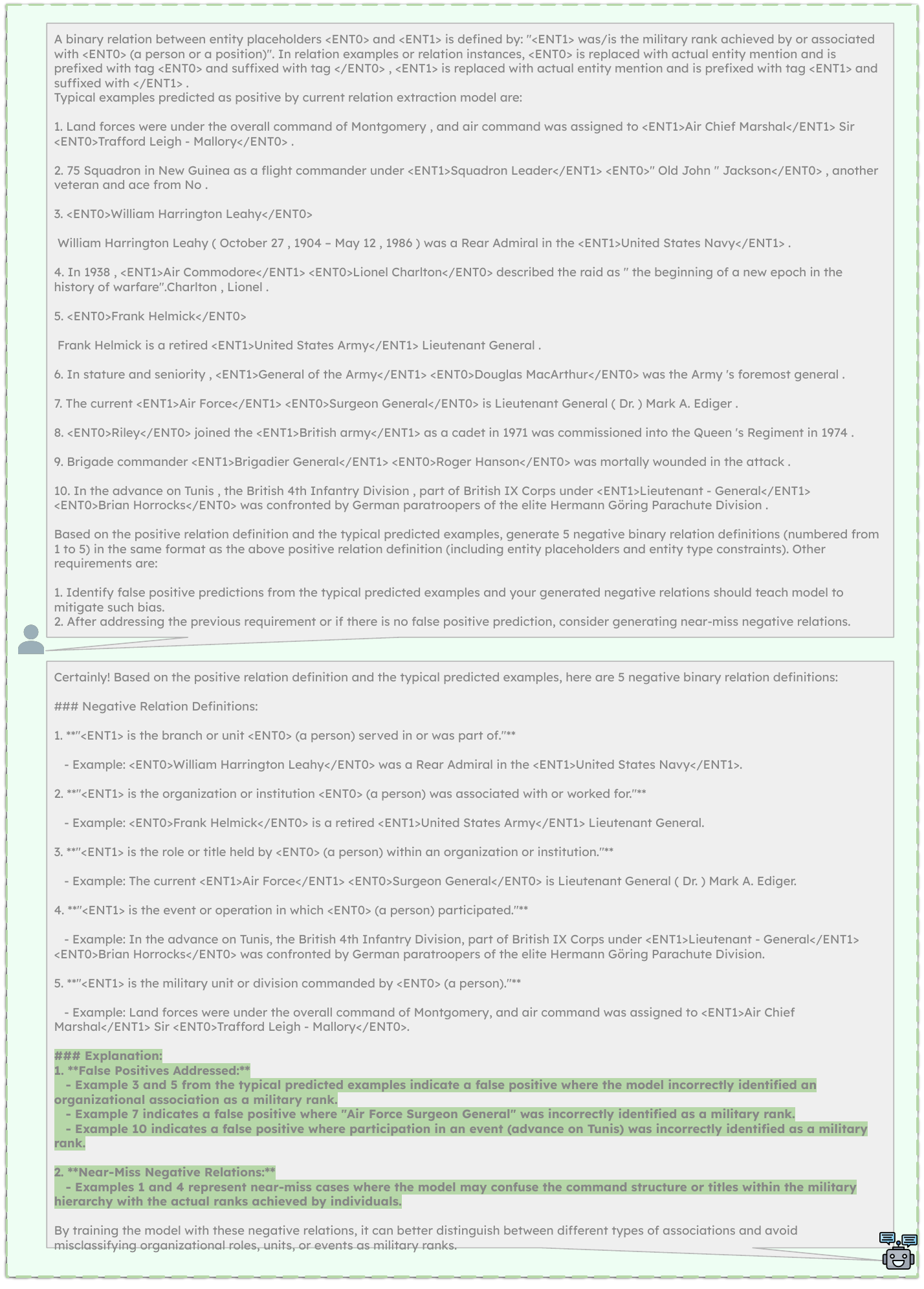}
    \caption{Example interaction dialogue which demonstrates the feedback-driven generation of negative relation definitions.}
    \label{fig:case study for neg def gen}
\end{figure*}

In order to showcase how LLMs function in accordance with our design expectations, we provide some dialogue cases in Fig.~\ref{fig:case study for pos gen} and Fig.~\ref{fig:case study for neg def gen}.
From the example dialogues, we can see that GPT-4o is considerably capable of identifying and summarizing the existing patterns in the dialogue history.  
The conversational feature also improves the interpretability of the results which can be further exploited with human-in-the-loop methods for future research or downstream applications. 

It's also evident that feedback-driven negative relation definition generation is of great potential. 
As the feedback instances may contain false predictions which can be directly taken by LLM to generate new negative relations and effectively rectify the SLM's bias.
Such feature directly boost the precision score as shown in Sec.~\ref{sec::results and analysis:ablation study}. 
Hence, we can see that our design expectations are well fulfilled and this also qualitatively explains the performance boost of our proposed model. 

Meanwhile, Fig.~\ref{fig:case study for neg def gen} also indicates that GPT-4o is still not perfect on giving the correct intermediate analysis of examining each of the feedback example. 
This indicates the future work can further improve the design of feedback acquisition and feedback processing. 

\end{document}